\documentclass[lettersize,journal]{IEEEtran}
\usepackage{xcolor}

\usepackage{multirow}
\usepackage{resizegather}
\usepackage{makecell, rotating}
\usepackage{array}
\usepackage{colortbl}

\ifCLASSOPTIONcompsoc
  \usepackage[nocompress]{cite}
\else
  \usepackage{cite}
\fi
\usepackage[hidelinks]{hyperref}

\usepackage{epsfig,graphicx}

\usepackage{amsmath,amssymb}
\allowdisplaybreaks

\usepackage{algorithm}
\usepackage{algpseudocode}

\ifCLASSOPTIONcompsoc
  \usepackage[caption=false,font=footnotesize,labelfont=sf,textfont=sf]{subfig}
\else
  \usepackage[caption=false,font=footnotesize]{subfig}
\fi

\usepackage{url}

\begin{document}

\title{Confidence-Aware Active Feedback for\\ Interactive Instance Search}

\author{Yue~Zhang,
       Chao~Liang,
       and~Longxiang~Jiang
\thanks{This work was supported in part by the National Natural Science Foundation of China under Grant U1903214, 61862015, 61876135, and 62071338, and in part by the Fundamental Research Founds for the Central Universities under Grant 2042022kf0001. \textit{(Corresponding author: Chao Liang.)}}
\thanks{The authors are with the National Engineering Research Center for Multimedia Software, School of Computer Science, Wuhan University, Wuhan 430072, China, and also with the Hubei Key Laboratory of Multimedia and Network Communication Engineering, Wuhan University, Wuhan 430072, China (e-mail: \{moozy924, cliang, jianglx\}@whu.edu.cn).}%
\thanks{This article has been accepted for publication in IEEE Transactions on Multimedia. This is the author's version which has not been fully edited and content may change prior to final publication. Citation information: DOI 10.1109/TMM.2022.3217965}
}

\maketitle

\IEEEdisplaynontitleabstractindextext


\begin{abstract}
  Online relevance feedback (RF) is widely utilized in instance search (INS) tasks to further refine imperfect ranking results, but it often has low interaction efficiency. The active learning (AL) technique addresses this problem by selecting valuable feedback candidates. However, mainstream AL methods require an initial labeled set for a cold start and are often computationally complex to solve. Therefore, they cannot fully satisfy the requirements for online RF in interactive INS tasks. To address this issue, we propose a confidence-aware active feedback method (CAAF) that is specifically designed for online RF in interactive INS tasks. Inspired by the explicit difficulty modeling scheme in self-paced learning, CAAF utilizes a pairwise manifold ranking loss to evaluate the ranking confidence of each unlabeled sample. The ranking confidence improves not only the interaction efficiency by indicating valuable feedback candidates but also the ranking quality by modulating the diffusion weights in manifold ranking. In addition, we design two acceleration strategies, an approximate optimization scheme and a top-$K$ search scheme, to reduce the computational complexity of CAAF. Extensive experiments on both image INS tasks and video INS tasks searching for buildings, landscapes, persons, and human behaviors demonstrate the effectiveness of the proposed method. Notably, in the real-world, large-scale video INS task of NIST TRECVID 2021, CAAF uses 25\% fewer feedback samples to achieve a performance that is nearly equivalent to the champion solution. Moreover, with the same number of feedback samples, CAAF's mAP is 51.9\%, significantly surpassing the champion solution by 5.9\%. Code is available at \url{https://github.com/nercms-mmap/caaf}.
  
\end{abstract}
  

\section{Introduction}\label{sec:introduction}

\IEEEPARstart{W}{ith} the explosive development of media technology, image/video instance analysis has received much attention in the past decade~\cite{TIP15-INS,TMM16-INS,TMM18-INS,TPAMI11-action,TIP20-action,TIP21-action,TIP21-hashing,TMM21-metric}. The goal of instance search (INS) is to search for instances of a query from a gallery of images or videos. After the features and initial ranking list are obtained, online relevance feedback (RF) is commonly employed in INS tasks to refine the imperfect retrieval result by asking for extra supervisory information through an interactive process~\cite{TMM18-RF,KBS21-RF}. However, RF often has low interaction efficiency, as not all samples are guaranteed to provide valuable information. 

Active learning (AL)~\cite{TR09-AL} addresses this issue by selecting valuable feedback samples that are beneficial for improving the model performance. Particularly, considering the practical difficulty of collecting labeled data and the special requirement of quick response in the online interaction scenario, we need a lightweight AL strategy that does not require a predefined labeled set and that has low computational complexity. However, mainstream AL methods, \textit{e.g.},~\cite{ICLR18-CoreSet,ICCV19-VAAL,AAAI19-SPAL,CVPR21-SequentialGCN}, require an initial labeled pool that accounts for 5\%$\sim$10\% of the entire training set for a cold start, and solving them is often time-consuming. Therefore, existing AL methods cannot fully satisfy the need for online RF in interactive INS tasks.

\begin{figure*}[!t]
  \centering
  \includegraphics[width = .9\linewidth]{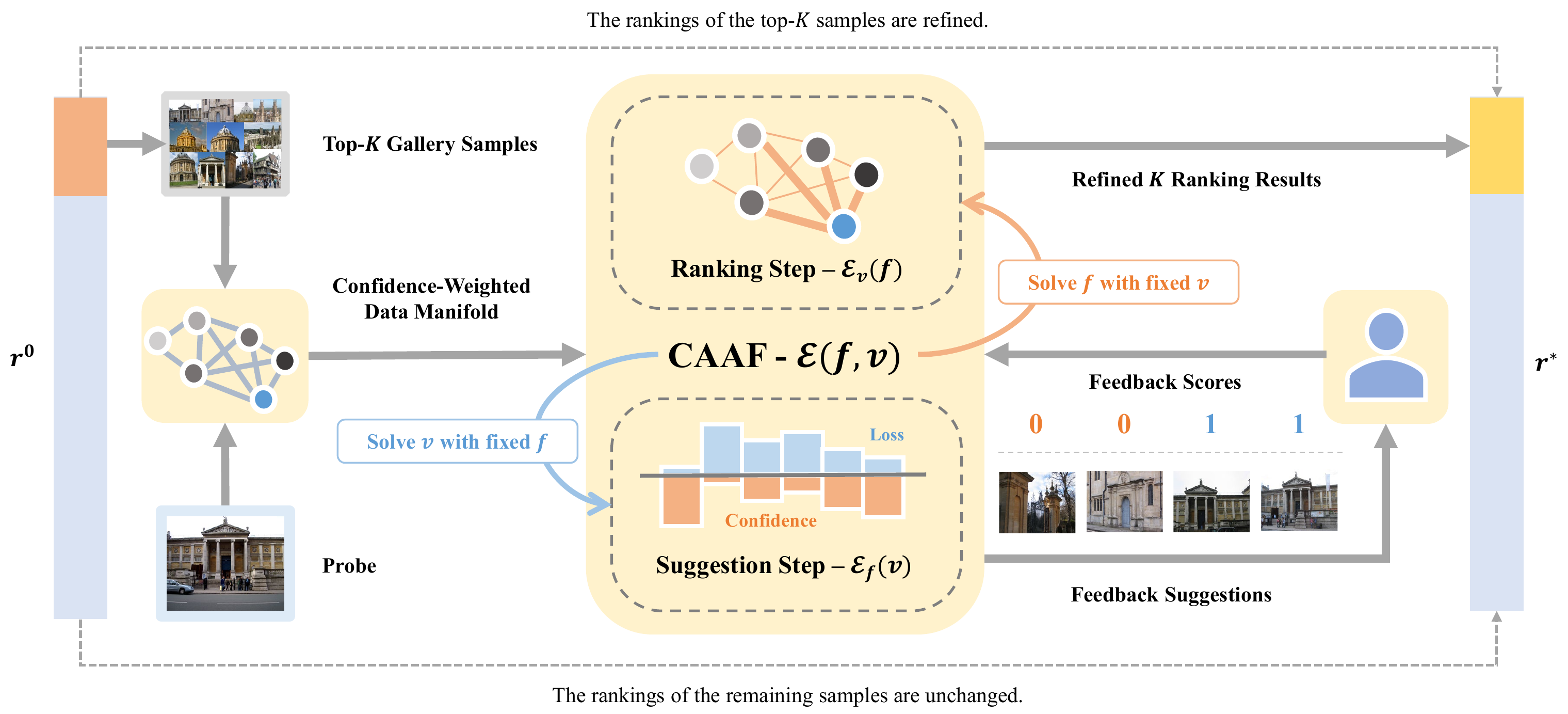}
  \caption{Workflow of the proposed CAAF. Given an initial ranking list $\boldsymbol{r}^0$, where gallery samples are listed in descending order by their initial ranking scores, the pairwise similarity between the top-ranked $K$ gallery samples and the probe is depicted by a confidence-weighted data manifold. On this basis, CAAF couples the ranking score $\boldsymbol{f}$ and confidence score $\boldsymbol{v}$ in one objective function $\mathcal{E}(\boldsymbol{f},\boldsymbol{v})$, where $\boldsymbol{f}$ is solved by minimizing $\mathcal{E}_{\boldsymbol{v}}(\boldsymbol{f})$ in the \textit{ranking step} and $\boldsymbol{v}$ is solved by minimizing $\mathcal{E}_{\boldsymbol{f}}(\boldsymbol{v})$ in the \textit{suggestion step}. Feedback suggestions are returned to the user, and the feedback scores are used to refine the ranking scores in the next round of feedback. The re-ranked top-$K$ samples, which are listed in descending order by their refined ranking scores (represented as a light yellow bar), are merged back to the initial ranking list (represented as a light blue bar), and the final ranking list $\boldsymbol{r}^*$ is generated.}
  \label{img-caaf}
\end{figure*}

To address the above issue, we propose a confidence-aware active feedback method (CAAF) that is specifically designed for online RF in interactive INS tasks. Inspired by the explicit difficulty modeling scheme in self-paced learning (SPL)~\cite{NIPS10-SPL,NIPS14-SPLD}, CAAF measures the confidence of each sample with the manifold ranking (MR)~\cite{NIPS03-MR} loss. Since MR does not require any labeled samples except for the query, CAAF is compatible with interactive INS tasks where the query is the only labeled sample at the beginning. However, unlike SPL, which focuses on easy samples with small losses, CAAF preferentially selects low-confidence hard samples with large losses, as they have higher potential values from the perspective of AL~\cite{TPAMI18-ASPL}.

To solve the proposed CAAF, we adopt the alternative optimization strategy~\cite{alt-opt}, which divides CAAF into a \textit{ranking step} and a \textit{suggestion step}. The \textit{ranking step} refines ranking scores with the user's relevance feedback, where labeled samples are endowed with higher confidence scores and have higher weights in diffusing their ranking scores. The \textit{suggestion step} estimates samples' confidence scores with their ranking losses, where unlabeled samples with larger ranking losses gain lower confidence scores. As a result, CAAF improves not only the interaction efficiency by selecting valuable feedback samples but also the retrieval performance by modulating the diffusion weights in MR.\IEEEpubidadjcol

However, the above solution consists of two quadratic programming problems, which still suffer from high computational complexity. Therefore, we further design two acceleration strategies to reduce the execution time of CAAF. We abandon the convergence condition of the alternative optimization strategy and approximately solve the quadratic-programming-based objective function by calculating the closed-form expressions, \textit{i.e.}, the \textit{ranking step} and the \textit{suggestion step} are successively solved only once in each round of feedback instead of iteratively optimized until convergence, and the constraints in their objective functions are omitted. On the other hand, only top-$K$ samples in the initial ranking list are considered in the interactive INS process. Hence, the problem scale is confined to the constant $K$ which is usually much smaller than the gallery size. The experimental results in Section~\ref{exp:rationality} demonstrate that both strategies significantly reduce the execution time without noticeable performance degradation.

The workflow of the proposed CAAF method is shown in Figure~\ref{img-caaf}. Given an initial ranking list, first, we select the top-$K$ gallery samples. Second, we construct a confidence-weighted data manifold with these samples and the probe to depict their pairwise relationship, where the nodes correspond to the probe and the galleries, and the edges encode their pairwise similarities. On this basis, CAAF successively solves a \textit{ranking step} and a \textit{suggestion step} to generate refined ranking results and feedback suggestions, respectively. This RF process will repeat several times until the maximum annotation budget is reached. Eventually, the final ranking list concatenates the re-ranked top-$K$ samples and the remaining samples in their initial ranking orders.

The contributions of this paper are summarized as follows:
\begin{itemize}
  \item We propose CAAF, an AL strategy that is specifically designed for interactive INS tasks, which improves not only the interaction efficiency by selecting valuable feedback samples but also the retrieval performance by modulating the diffusion weights in manifold ranking.
  \item We design two acceleration strategies to solve CAAF, including an approximate solution to simplify the solving process and a top-$K$ search scheme to reduce the problem scale, which ensure a smooth interaction experience in large-scale INS tasks.
  \item We conduct extensive experiments on both image and video INS tasks searching for buildings, landscapes, persons, and human behaviors, which demonstrate the effectiveness of the proposed method. Notably, in the large-scale video INS task of NIST TRECVID 2021, CAAF uses 25\% fewer feedback samples to achieve nearly equivalent performance as the champion solution. Moreover, with the same number of feedback samples, CAAF's mAP is 51.9\%, significantly surpassing the champion solution by 5.9\%.
\end{itemize}

The remainder of this paper is organized as follows: Section~\ref{sec:related-work} briefly reviews the related work. In Section~\ref{sec:proposed-method}, we describe the proposed CAAF method and its solution. Section~\ref{sec:experiments} shows the experimental results, and conclusions are given in Section~\ref{sec:conclusions}. 

\section{Related Work}\label{sec:related-work}

In this section, we briefly present a review of INS and RF and then introduce related developments in AL and SPL.

\subsection{Instance Search}
INS tasks aim to search a specific instance from a large set of multimedia data. According to the data type, INS tasks include image retrieval~\cite{TPAMI18-INS-Survey} and video retrieval~\cite{TMM21-videoretrieval}. According to the category of the search instance, INS tasks include but are not limited to object retrieval~\cite{TMM18-INS} and person re-identification~\cite{TMM22-reid}. A typical pipeline of the INS task is to detect and locate the search instance~\cite{TMM16-INS,TPAMI11-action,TIP20-action,TIP21-action}, to extract and encode discriminative features to compact vectors~\cite{TIP21-hashing,TIP15-INS,TMM18-INS,TMM21-metric}, and then to apply post-processing strategies to improve the retrieval accuracy~\cite{TMM19-QE,Access21-RF}. 

Our method is a post-processing strategy that aims to refine the existing retrieval results through efficient human interactions. The method has been tested to be effective on image- and video-based datasets covering buildings~\cite{CVPR07-Oxford5k}, landscapes~\cite{ECCV08-Holidays}, persons~\cite{CVPR14-CUHK03}, and human behaviors~\cite{techreport}. 

\subsection{Relevance Feedback}
RF is a commonly utilized post-processing technique that refines the existing retrieval results by asking users to provide extra annotations. Generally, RF comprises two steps: sample selection and ranking optimization. Existing methods mainly focus on the second procedure, \textit{i.e.}, optimizing the model with a small amount of user feedback to generate more accurate retrieval results~\cite{TMM18-RF,KBS21-RF}. However, the first procedure is naively implemented by top selection or random selection, which is not guaranteed to select the most valuable feedback candidates, thus leading to low interaction efficiency~\cite{MTA17-AL4Retrieval}. 

To address this problem, some researchers have introduced AL techniques to RF~\cite{MTA17-AL4Retrieval}. CAAF follows this rationale, but implements the idea differently. The details are described in the following sections.

\subsection{Active Learning}
AL methods aim to minimize the annotation cost by selecting the most valuable feedback samples that are either informative~\cite{ICMLA18-DeepBASS,ICCV19-DRAL}, representative~\cite{ICLR18-CoreSet,ICCV19-VAAL}, or both~\cite{CVPR12-RALF,CVPR21-SequentialGCN}. The mainstream AL methods require a pre-defined labeled set for a cold start, \textit{e.g.}, VAAL~\cite{ICCV19-VAAL} initializes the labeled pool with 10\% of the training set, and CoreGCN~\cite{CVPR21-SequentialGCN} randomly selects 1,000 samples as the initial labeled set. In addition, these methods are often computationally complex to solve, \textit{e.g.}, VAAL needs to train a variational autoencoder and an adversarial network to select representative unlabeled samples, and CoreGCN estimates the uncertainty of unlabeled samples by training a graph convolutional network. However, considering the practical difficulty of collecting labeled data and the special requirement of quick response in the online interaction scenario, these methods cannot fully satisfy the need for online RF in interactive INS tasks.

CAAF addresses this problem by measuring the samples' confidence with the manifold ranking~\cite{NIPS03-MR} loss and preferentially selects low-confidence hard samples with large losses as valuable feedback candidates. Since MR does not require any labeled samples except for the query, CAAF is compatible with interactive INS tasks where the query is the only labeled sample at the beginning.

\subsection{Self-Paced Learning}
SPL~\cite{NIPS10-SPL,NIPS14-SPLD} is a classic learning paradigm that gradually incorporates easy to hard samples into the training processes. The core idea of SPL is to measure the easiness of each sample with a task-specific loss and to preferentially select easy samples with low losses to train the model. The effectiveness of such a learning regime has been validated in various tasks, including fine-grained visual classification~\cite{TIP21-SPL-arcface}, cross-modal matching~\cite{ACMMM21-SPL-triplet} and object detection~\cite{IJCV19-SPL,TPAMI20-SPL,NIPS20-SPL}.

Recently, some methods~\cite{TPAMI18-ASPL,AAAI19-SPAL} have combined SPL and AL into one framework, \textit{e.g.}, ASPL~\cite{TPAMI18-ASPL} assigns pseudo labels for high-confidence easy samples selected by SPL and annotates ground-truth labels for low-confidence samples selected by AL; and SPAL~\cite{AAAI19-SPAL} integrates SPL and AL into one objective function, so the easiness and representativeness of each sample can be jointly considered for model training.

Different from ASPL and SPAL, CAAF directly uses SPL as an AL strategy and preferentially selects low-confidence hard samples with large losses as valuable candidates for online RF in interactive INS tasks. In addition, CAAF uses SPL as a modulator to adjust the diffusion weights in MR, where labeled samples are endowed with higher confidence scores and thus have higher weights in diffusing their ranking scores. As a result, CAAF improves not only the interaction efficiency but also the ranking accuracy.

\section{Proposed Method}\label{sec:proposed-method}

In this section, we present the problem formulation with necessary notations and then introduce the solution and acceleration strategies.

\subsection{Problem Formulation}

Given a probe $p$ and a gallery set $\mathcal{G} =\left\{g_i\right\}_{i=1}^n$, we merge them into a new image set $\mathcal{X} = \mathcal{G}\cup\left\{p\right\} = \left\{x_i\right\}_{i=1}^{m}$, where $m=n+1$. Next, the pairwise data affinity can be represented as $\mathbf{A} = [\max(a_{ij},0)]_{i,j=1}^{m}\in[0,1]^{m\times m}$, where $a_{ij}$ reflects the similarity between $x_i$ and $x_j$. Our goal is to actively select the most valuable feedback samples for RF to refine the ranking score $\boldsymbol{f} = [f_i]_{i=1}^{m}\in[0,1]^{m\times1}$, where $f_i$ is the ranking score of $x_i$.
  
\textbf{Manifold ranking. }The classic manifold ranking (MR)~\cite{NIPS03-MR} aims to obtain the ranking score $\boldsymbol{f}$ by minimizing a pairwise ranking loss function, with a smoothing term to constrain similar samples to have similar ranking scores, and a fitting term to prevent the ranking score $\boldsymbol{f}$ from deviating too much from the reference ranking score $\boldsymbol{y} = [0, \cdots, 0, 1]^{\top}\in\left\{0,1\right\}^{m\times1}$. The pairwise loss function between $x_i$ and $x_j$ is defined as
\begin{eqnarray}
  l_{ij} = a_{ij} (f_i - f_j)^2 + \alpha(f_i - y_i)^2 + \alpha(f_j - y_j)^2,  \label{eq:orimanif}
\end{eqnarray}
where $\alpha\in(0,1)$ is a parameter to balance the smoothing term and fitting term. To adapt such a framework to our human-in-the-loop setting, we redefine $\boldsymbol{y}=[y_i]_{i=1}^{m}$ as
\begin{eqnarray}
 {y_i} = \left\{
  \begin{aligned}
    s_i, & \text{\quad if }x_i\in\Psi, \\
    0, & \text{\quad otherwise},
  \end{aligned}\right.
  \label{eq:y0}
\end{eqnarray}
where $s_i\in\{0,1\}$ is the feedback score of $x_i$, and $\Psi$ is the set of all annotated samples. $s_i=1$ if $x_i$ is relevant to the probe, and $s_i=0$ otherwise. In particular, probe $x_m=p$ can be considered a special annotated sample whose feedback score is $s_{m}=1$. $f_i=y_i$ if $x_i\in\Psi$. 

\textbf{Confidence modeling. }In the context of INS, the confidence of a sample can be defined as the reliability of its ranking score. We denote the confidence score as $\boldsymbol{v} = [v_i]_{i=1}^{m}\in[0,1]^{m\times1}$, where $v_i$ is the confidence score of $x_i$. Apparently, the labeled samples have higher confidence than the unlabeled samples. Hence, $v_i$ is initialized as
\begin{eqnarray}
 {v_i} = \left\{
  \begin{aligned}
    1, & \text{\quad if }x_i\in\Psi, \\
    0, & \text{\quad otherwise}.
  \end{aligned} \right.
  \label{eq:v0}
\end{eqnarray} 
For unlabeled samples, however, $v_i$ is negatively correlated with the ranking loss. \textit{i.e.}, the higher the ranking loss is, the lower the confidence. And from the perspective of AL, the lower the confidence is, the higher the feedback value. Therefore, $\boldsymbol{v}$ is optimized by minimizing the following loss function:
\begin{eqnarray}
  \mathcal{L}(\boldsymbol{f},\boldsymbol{v}) = \frac{1}{m^2}\sum_{i,j}(v_i+v_j)(l_{ij}-\beta), \label{eq:lossmain}
\end{eqnarray}
where $l_{ij}$ is the pairwise manifold ranking loss defined in Eq.~\ref{eq:orimanif}, and $\beta>0$ is a loss threshold to measure the confidence of each sample. Specifically, when $l_{ij}$ is smaller than $\beta$, minimizing $\mathcal{L}(\boldsymbol{f},\boldsymbol{v})$ will make both $v_i$ and $v_j$ approach 1; in contrast, when $l_{ij}$ is larger than $\beta$, minimizing $\mathcal{L}(\boldsymbol{f},\boldsymbol{v})$ will make both $v_i$ and $v_j$ approach 0. 

By optimizing Eq.~\ref{eq:lossmain}, $\boldsymbol{v}$ works in two aspects: \label{sec:cwmr} 
\begin{itemize}
  \item \textbf{Feedback generation. }When taking $\boldsymbol{v}$ itself as the optimization objective, it serves as an indicator to generate feedback suggestions based on the ranking confidence of each unlabeled sample. 
  \item \textbf{Weight modulation. }When taking $\boldsymbol{f}$ as the optimization objective, $\boldsymbol{v}$ becomes an additional weight in MR to increase the impact of high-confidence samples and to reduce the influence of the low-confidence samples.
\end{itemize}

Furthermore, to constrain the element value of $\boldsymbol{v}$, a squared norm regularization term $\mathcal{R}(\boldsymbol{v})$ is added to Eq.~\eqref{eq:lossmain}.
\begin{eqnarray}
  \mathcal{R}(\boldsymbol{v}) = \frac{\gamma}{m}\left \|  \boldsymbol{v}\right \| _{2}^{2}, 
\end{eqnarray}
where $\gamma > 0$ is the regularization weight parameter. We obtain the following constraint optimization problem:
\begin{eqnarray}
  \begin{aligned}
    \min_{\boldsymbol{f}, \boldsymbol{v}}\text{ } & \mathcal{E}(\boldsymbol{f}, \boldsymbol{v})=\mathcal{L}(\boldsymbol{f}, \boldsymbol{v})+\mathcal{R}(\boldsymbol{v}) \\ 
    \text{s.t. } & \mathbf{0} \preceq \boldsymbol{f} \preceq \mathbf{1}\text{, } f_i=y_i\text{ if }x_i\in\Psi \\
    ~ & \mathbf{0} \preceq \boldsymbol{v} \preceq \mathbf{1}\text{, } v_i=1\text{ if }x_i\in\Psi
  \end{aligned} \label{eq:ojectmain} 
\end{eqnarray}
where $\mathbf{0}$ and $\mathbf{1}$ represent a full $0$ vector and a full $1$ vector, respectively, and $\preceq$ denotes the element-wise comparison of $\le$.

\subsection{Solution}\label{sec:solution}
To solve $\boldsymbol{f}$ and $\boldsymbol{v}$, we adopt the alternative optimization strategy~\cite{alt-opt} and decompose Eq.~\ref{eq:ojectmain} into a \textit{ranking step} and \textit{suggestion step}.

\textbf{Ranking step. } In the ranking step, we optimize $\boldsymbol{f}$ with fixed $\boldsymbol{v}$; thus, Eq.~\eqref{eq:ojectmain} can be simplified by eliminating the regularization term $\mathcal{R}(\boldsymbol{v})$ and the threshold $\beta$. The objective function is then computed as 
\begin{eqnarray}
  \mathcal{E}_{\boldsymbol{v}}(\boldsymbol{f}) = \frac{1}{2}\sum_{i,j}\tilde{v}_{ij}l_{ij},\label{eq:objectf}
\end{eqnarray}
where $\tilde{v}_{ij} = v_i+v_j$ and the constant coefficient $\frac{1}{m^2}$ in Eq.~\eqref{eq:lossmain} is replaced by $\frac{1}{2}$ to simplify the following transformation. By replacing $l_{ij}$ with Eq.~\eqref{eq:orimanif} and by eliminating terms irrelevant to $\boldsymbol{f}$, Eq.~\eqref{eq:objectf} can be further rewritten as a confidence-weighted manifold ranking problem \label{sec:ranking}
  \begin{align}
    \mathcal{E}_{\boldsymbol{v}}(\boldsymbol{f}) =& \frac{1}{2}\sum_{i,j}\tilde{v}_{ij}l_{ij}\notag  \\
    =& \frac{1}{2}\sum_{i,j}\tilde{v}_{ij}a_{ij} (f_i - f_j)^2 + \sum_{i,j}\alpha\tilde{v}_{ij}(f_i - y_i)^2\notag  \\
    =& \frac{1}{2}\sum_{i,j}\tilde{v}_{ij}a_{ij} (f_i^2+f_j^2-2f_if_j) +\notag  \\
    &  \sum_{i,j}\alpha\tilde{v}_{ij}(f_i^2 + y_i^2 - 2f_iy_i) \label{eq:objectfmatrix}  \\
    =& \sum_{i,j} \tilde{a}_{ij}{f_i}^2 - \sum_{i,j} \tilde{a}_{ij}f_if_j + \sum_{i,j}\alpha\tilde{v}_{ij}f_i^2 - \notag \\
    &   2\sum_{i,j}\alpha\tilde{v}_{ij}f_iy_i + \sum_{i,j}\alpha\tilde{v}_{ij}y_i^2 \notag \\
    =& \boldsymbol{f}^\top\mathbf{D}\boldsymbol{f}-\boldsymbol{f}^\top \mathbf{\tilde{A}} \boldsymbol{f} + \boldsymbol{f}^\top\mathbf{Q}\boldsymbol{f}-2\boldsymbol{f}^\top\mathbf{Q}\boldsymbol{y}+C \notag \\
    =& \boldsymbol{f}^{\top}(\mathbf{P}+\mathbf{Q})\boldsymbol{f}-2\boldsymbol{f}^{\top}\mathbf{Q}\boldsymbol{y}+C, \notag
  \end{align}
where $\tilde{a}_{ij}=\tilde{v}_{ij}a_{ij}$ is the confidence-weighted affinity between $x_i$ and $x_j$; $\mathbf{D}$ is a diagonal matrix, where $d_{ii} = \sum_{j}\tilde{a}_{ij}$; $\mathbf{Q}$ is a diagonal matrix, where $q_{ii} = \sum_{j}\alpha\tilde{v}_{ij}$; $\mathbf{P}=\mathbf{D}-\mathbf{\tilde{A}}$ is the Laplacian matrix of the confidence-weighted graph; and $C = \sum_{i,j}\alpha\tilde{v}_{ij}y_i^2$ is a constant term irrelevant to $\boldsymbol{f}$. Since both $\mathbf{P}$ and $\mathbf{Q}$ are positive semi-definite matrices, Eq.~\eqref{eq:objectfmatrix} is a standard convex quadratic programming (QP) problem~\cite{convex-opt}. 

\textbf{Suggestion step. }In the suggestion step, $\boldsymbol{f}$ is fixed while $\boldsymbol{v}$ is optimized. The related objective function can be concisely rewritten in a matrix form as
\begin{eqnarray}
  \begin{aligned}
    \mathcal{E}_{\boldsymbol{f}}(\boldsymbol{v}) 
     =& \frac{\gamma}{m}\left \|  \boldsymbol{v}\right \| _{2}^{2}  + \frac{1}{m^2}\sum_{i,j}(v_i+v_j)(l_{ij}-\beta) \\
     =& \frac{\gamma}{m}\sum_{i}v_i^2 + \frac{2}{m^2}\sum_{i}v_{i}\tilde{l}_i \\
     =& \frac{\gamma}{m}\boldsymbol{v}^{\top}\mathbf{I}\boldsymbol{v}+\frac{2}{m^2}\tilde{\boldsymbol{l}}^{\top}\boldsymbol{v} 
  \end{aligned} \label{eq:objectv}
\end{eqnarray}
where $\mathbf{I}$ is an $m$-by-$m$ unit matrix, $\tilde{\boldsymbol{l}} = [\tilde{l}_i]_{i = 1}^{m}\in\mathbb{R} ^{m\times 1}$, and $\tilde{l}_i = \sum_{j}(l_{ij}-\beta)$. Similarly, Eq.~\eqref{eq:objectv} is also a standard QP problem.

After the above decomposition, CAAF can generate one round of refined ranking scores and valuable feedback suggestions by alternatively optimizing the \textit{ranking step} and \textit{suggestion step} until convergence.

\subsection{Acceleration Strategies}\label{sec:implementation}

Real-time response is in high demand in interactive INS tasks, yet it may be too time-consuming to iteratively solve the QP problems in the \textit{ranking step} and \textit{suggestion step}, especially when the problem scale $m$ is extremely large. Therefore, we design two acceleration strategies to efficiently solve the proposed model, including an approximate solution to simplify the original solution in Section~\ref{sec:solution}, and a top-$K$ search scheme to reduce the problem scale.

\textbf{Approximate solution.}~To reduce the execution time, we abandon the convergence condition in the alternative optimization strategy and successively solve the \textit{ranking step} and \textit{suggestion step} only once in each round of feedback. In this case, we do not have to strictly follow the constraints in Eq.~\eqref{eq:objectfmatrix} and Eq.~\eqref{eq:objectv}, and both $\boldsymbol{f}$ and $\boldsymbol{v}$ can be solved by closed-form expressions.

In the \textit{ranking step}, $\boldsymbol{f}$ can be solved by
\begin{eqnarray}
    f_i = \left\{
      \begin{aligned}
        & \text{\quad\quad\quad} y_i & , & \text{\quad if }x_i\in\Psi, \\
        & \frac{f_i-\min(\boldsymbol{\hat{f}})}{\max(\boldsymbol{\hat{f}})-\min(\boldsymbol{\hat{f}})} & , & \text{\quad otherwise},
      \end{aligned} \right.
     \label{eq:resultf} 
\end{eqnarray}
where $\min(\cdot)$ and $\max(\cdot)$ denote the minimum and maximum value, respectively, of a vector, and $\boldsymbol{\hat{f}}$ is solved by
\begin{eqnarray}
  \boldsymbol{\hat{f}} = (\mathbf{P}+\mathbf{Q})^{-1}\mathbf{Q}\boldsymbol{y}.\label{eq:resultf-1} 
\end{eqnarray}

In the \textit{suggestion step}, $\boldsymbol{v}$ can be solved by 
\begin{eqnarray}
  \boldsymbol{v} = -\frac{\tilde{\boldsymbol{l}}}{\gamma m} \label{eq:resultv-1}.
\end{eqnarray}
Since we are only concerned with the relative magnitude of $\boldsymbol{v}$ instead of its real value, the constant terms can be further omitted. Hence, $\boldsymbol{v}$ is solved by
\begin{eqnarray}
    \boldsymbol{v} = -\hat{\boldsymbol{l}}, \label{eq:resultv-2}
\end{eqnarray}
where $\hat{\boldsymbol{l}}=[\hat{l_i}]_{i=1}^n$ and $\hat{l_i}=\sum_j{l_{ij}}$.

Since Eq.~\eqref{eq:resultf} and \eqref{eq:resultv-2} only involve basic matrix operations, we can use a GPU to accelerate the computational process. The comparative results between the original QP solution and the GPU-accelerated approximate solution will be discussed in Section~\ref{exp:rationality}.

\begin{algorithm}[!t]  
  \caption{Approximate solution for CAAF}
  \label{alg:framwork}  
  \begin{algorithmic}[1] 
    \Require the image set $\mathcal{X} = \mathcal{G}\cup\left\{p\right\}$, where $\mathcal{G}$ and $p$ denote the gallery set and probe, respectively, the affinity matrix $\mathbf{A}$, the initial labeled image set $\Psi = \lbrace p \rbrace$, the number of feedback samples per round $q$, and the maximum round of feedback $T$.
    \Ensure The ranking score $\boldsymbol{f^*}$.
    \For{$t$ in $[0,T]$}
      \State Set $\boldsymbol{y}$ and $\boldsymbol{v}$ by Eq.\eqref{eq:y0} and Eq.\eqref{eq:v0}, respectively; 
      \State Solve $\boldsymbol{f}$ with fixed $\boldsymbol{v}$ by Eq.\eqref{eq:resultf}; 
      \State Solve $\boldsymbol{v}$ with fixed $\boldsymbol{f}$ by Eq.\eqref{eq:resultv-2};
      \State Select $q$ samples with the smallest $\boldsymbol{v}$ from $\mathcal{X}\setminus\Psi$, and add them to $\Psi$ with their feedback scores;
    \EndFor  
    \State \Return $\boldsymbol{f^*}=[f_i]_{i=1}^{n}$.
  \end{algorithmic}  
\end{algorithm}

After successively solving $\boldsymbol{f}$ and $\boldsymbol{v}$, the user is asked to provide feedback scores for the $q$ least confident unlabeled samples. At the beginning of the next round of feedback, we reuse Eq.~\eqref{eq:v0} to update $\boldsymbol{v}$, \textit{i.e.}, unlabeled samples are considered to be low-confidence while labeled samples are regarded as high-confidence. Once the pre-determined maximum round of feedback $T$ is reached, the ranking score of probe $f_m$ is removed, and the ranking score is denoted as $\boldsymbol{f^*}=[f_i]_{i=1}^{n}$. The overall procedure of the approximate solution is summarized in Algorithm~\ref{alg:framwork}. 

\textbf{Top-$K$ search scheme.}~Since relevant samples tend to be concentrated at the top of the initial ranking list~\cite{SIGIR98-why-top-k}, we follow a common practice that selects only the top-$K$ samples from the initial ranking list $\boldsymbol{r}^0$ to form the gallery set $\mathcal{G}$~\cite{SIGIR07-LOD,PRCV18-reid}, \textit{i.e.}, the number of galleries $n=K$, and the total number of samples $m=K+1$. Therefore, we can set a relatively small $K$ to balance the performance and computational cost. Eventually, the final ranking list $\boldsymbol{r}^*$ concatenates two parts: (1) the re-ranked top-$K$ samples listed in descending order by their refined ranking scores $\boldsymbol{f^*}$ and (2) the remaining samples in their initial ranking orders. The rationality of this practice will also be discussed in Section~\ref{exp:rationality}.

\section{Experiments}\label{sec:experiments}

This section is divided into six parts. In the first part, we introduce the settings in our experiments. Next, we report the comparative results with existing AL methods on three image-based INS datasets in the second part. The evaluation results on NIST TRECVID are reported in the third part. In the fourth part, ablation studies are performed to demonstrate the effectiveness of the two key modules in CAAF. In the fifth part, we provide a detailed analysis of the confidence modeling scheme introduced in Section~\ref{sec:cwmr} to show how $\boldsymbol{v}$ works in CAAF. The sixth part analyzes the rationality of the proposed acceleration strategies. We discuss the limitations of CAAF in the last part.

\subsection{Settings}
\label{sec:settings}
\subsubsection{Datasets}
We evaluate our method on three image retrieval datasets and the INS task of NIST TRECVID\footnote{\url{https://www-nlpir.nist.gov/projects/tv2021/ins.html}}: 

\begin{figure*}[t]
  \centering
  \includegraphics[width = \linewidth]{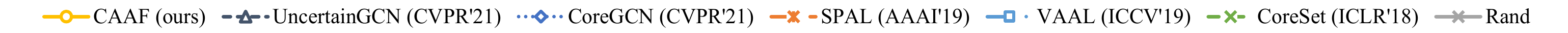}
  \subfloat[Holidays]{\includegraphics[width = 0.33\linewidth]{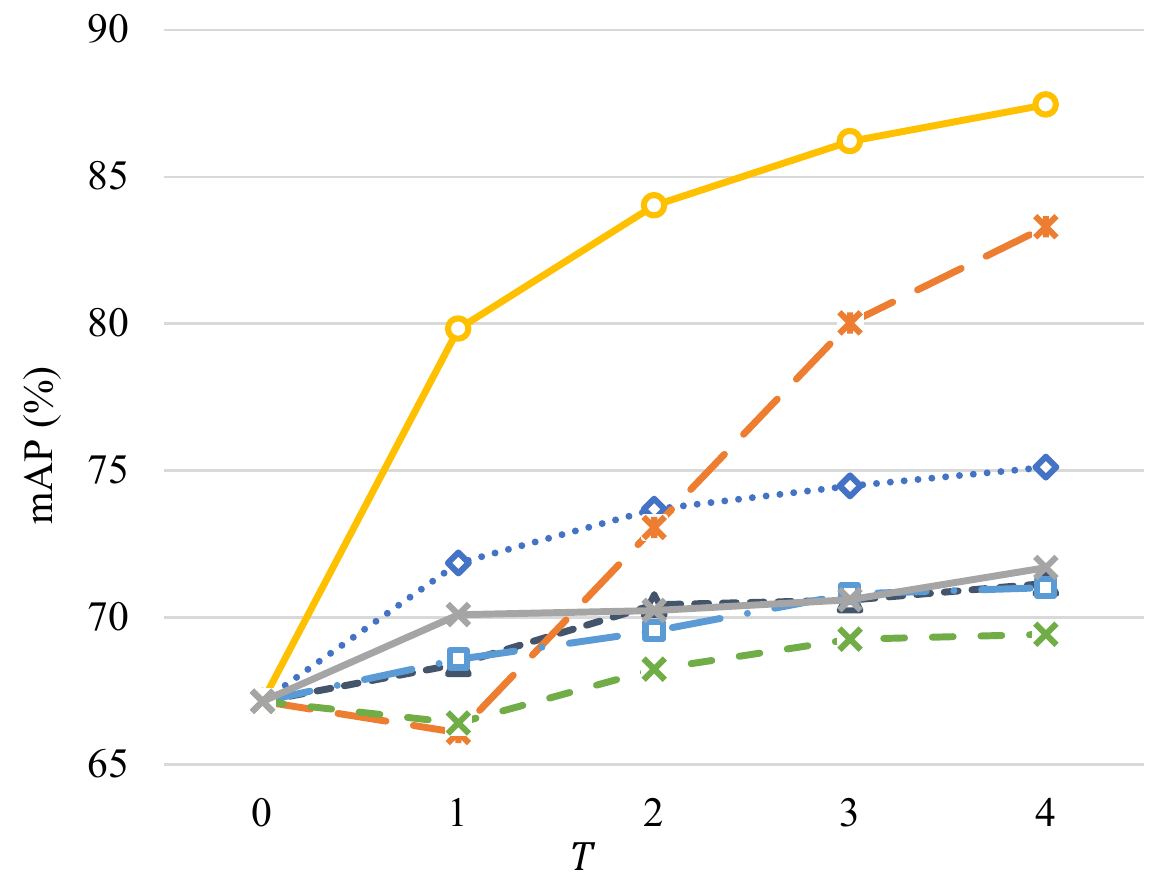}%
  \label{img-holiday-main}}
  \hfil
  \subfloat[Oxford5k]{\includegraphics[width = 0.33\linewidth]{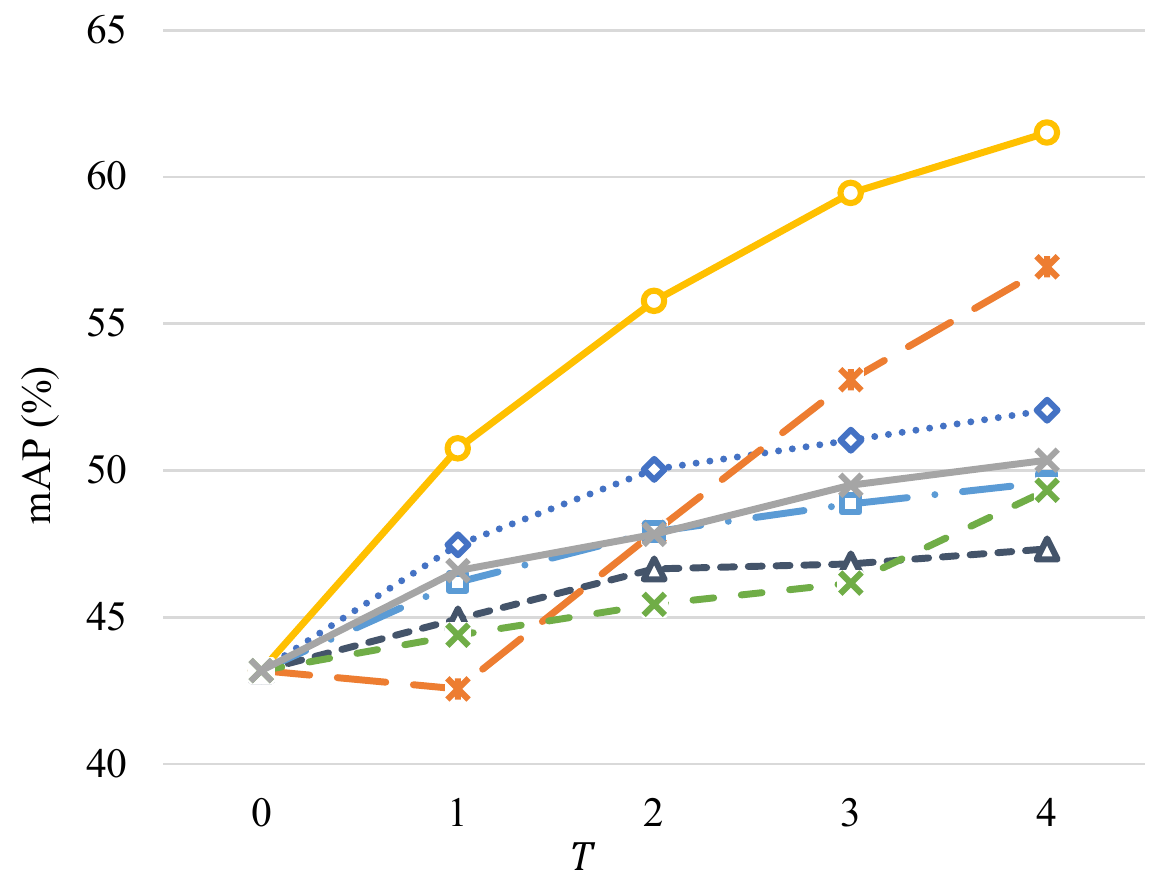}%
  \label{img-oxford-main}}
  \hfil
  \subfloat[CUHK03]{\includegraphics[width = 0.33\linewidth]{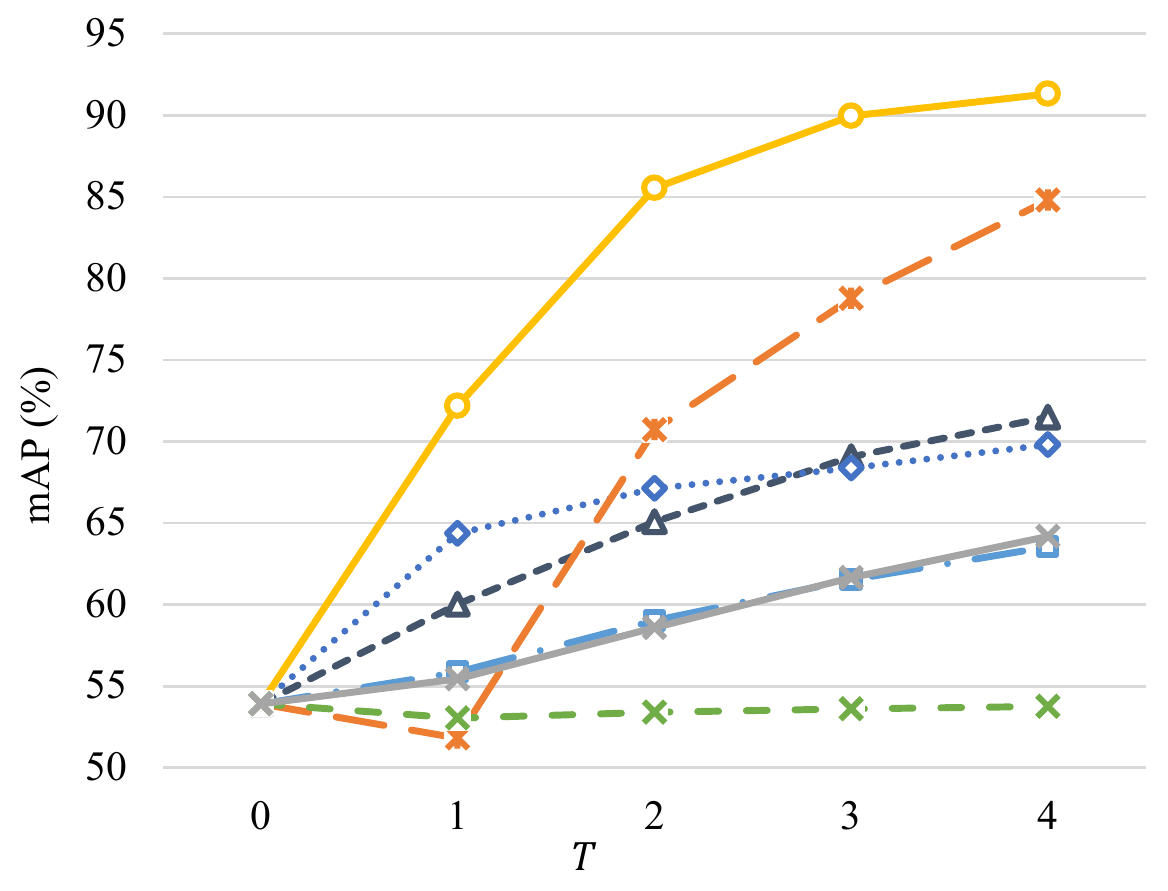}
  \label{img-cuhk03-main}}
  \hfil
  \caption{Performance comparison of different AL methods on the Holidays, Oxford5k, and CUHK03 datasets under different feedback rounds ($T$), where each line represents an AL method.}
  \label{img-exp-main}
\end{figure*}

\begin{itemize}
  \item \textbf{Holidays}~\cite{ECCV08-Holidays} is one of the most widely employed image retrieval benchmarks. There are 1,491 images of 500 landscapes collected from personal holiday albums, and each landscape has one query.
  \item \textbf{Oxford5k}~\cite{CVPR07-Oxford5k} is also a popular dataset for image retrieval. This dataset contains 5,062 images of 11 buildings in Oxford University, and 5 queries are defined for each building, \textit{i.e.}, there are 55 queries in total. 
  \item \textbf{CUHK03}~\cite{CVPR14-CUHK03} is a person re-identification dataset that collects images of 1,467 pedestrians on the CUHK campus. It provides two types of annotations: the first annotation is obtained by manually labeled bounding boxes (labeled set), and the second annotation is obtained by bounding boxes produced by an automatic detector (detected set). We choose the latter in our experiment, as it is more challenging than the former. We follow the new training/testing protocol introduced by Zheng et al.~\cite{CVPR17-k-reciprocal}, where there are 1,400 probes vs. 5,332 galleries in the testing set.
  \item \textbf{NIST TRECVID} INS Evaluation asks participants to retrieve all relevant video segments about specific topics from the \textit{EastEnders} series. Each participant can submit up to 4 automatic results and 4 interactive results. When generating the interactive results, human interactions are allowed to refine the automatically generated search results, and the interaction time for each search topic is limited to 5 minutes. The entire dataset contains more than 470,000 video shots from 244 episodes, with a total length greater than 464 hours. We choose 40 query topics about specific persons doing specific actions from the 2020 and 2021 challenges.
\end{itemize}

\subsubsection{Features}
We choose the 4,096-dimensional feature extracted by VGG16~\cite{ICLR15-VGG} pretrained on ImageNet~\cite{CVPR09-ImageNet} for Holidays and Oxford5k, and the 1,536-dimensional feature extracted by Beyond Part Model~\cite{ECCV18-PCB} is adopted for CUHK03. The affinity matrix $\mathbf{A}$ is constructed by cosine similarity, and the initial ranking list $\boldsymbol{r}^0$ is generated by sorting all gallery samples in descending order by their Euclidean distance to the probe. 

For NIST TRECVID, we concatenate the facial visual feature of the character that appeared in each video segment and the category encoding of action taken by the character as the semantic feature of each video segment~\cite{techreport}. We then take the average feature of the top 25 samples in the initial ranking list as the probe feature, as there is no specific query sample in the INS task of NIST TRECVID. Similar to the three image-based datasets, the affinity matrix $\mathbf{A}$ is calculated by the cosine similarity. However, we further add a temporal expansion term to reflect the temporal similarity between each video segment, \textit{i.e.}, $a_{ij} = e^{-\lambda|t_i-t_j|}\text{cos}(x_i,x_j)$, where $|t_i-t_j|$ calculates the temporal distance between shot $i$ and shot $j$, and $\text{cos}(x_i,x_j)$ denotes the cosine similarity between $x_i$ and $x_j$. The intuition is that temporally consecutive video segments tend to share the same topic, and we set $\lambda = 0.005$. Instead of calculating the Euclidean distance, the initial ranking list on NIST TRECVID is generated by an automatic ranking strategy~\cite{auto}.

\subsubsection{Evaluation metrics}
The mean average precision (mAP) is mainly used to evaluate the performance on all four datasets. We also measure the execution time taken in each round of feedback to evaluate the execution efficiency\footnote{The execution time of all experiments is measured on the same machine with one Intel(R) Core(TM) CPU i9-11900K @ 3.50 GHz, 128 GB memory, and one NVIDIA GeForce RTX 3090 GPU.}.

\subsubsection{Implementation details}\label{sec:params}
We set $\alpha=0.01$ on all datasets to balance the smoothing term and the fitting term in Equation~\ref{eq:orimanif}. On the three image-based datasets, we set $K=300$, $q=5$, and $T=4$, \textit{i.e.}, the top 300 samples in the initial ranking list are used to refine the retrieval performance, and $q \times T = 20$ feedback samples are generated for each probe in total. On NIST TRECVID, unless otherwise specified, we set $K=2000$ and $q=18$, but $T$ is no longer fixed since the total time is fixed in the official evaluation, and we try to take as many feedback rounds as possible. In addition, we replace the binary reference score $\boldsymbol{y}$ with the initial ranking score and set $\boldsymbol{v} = \boldsymbol{1}$ in the first round of feedback on NIST TRECVID, as the semantic features are somewhat weak to provide adequate information.

\begin{figure}[!t]
  \centering
  \includegraphics[width = .9\linewidth]{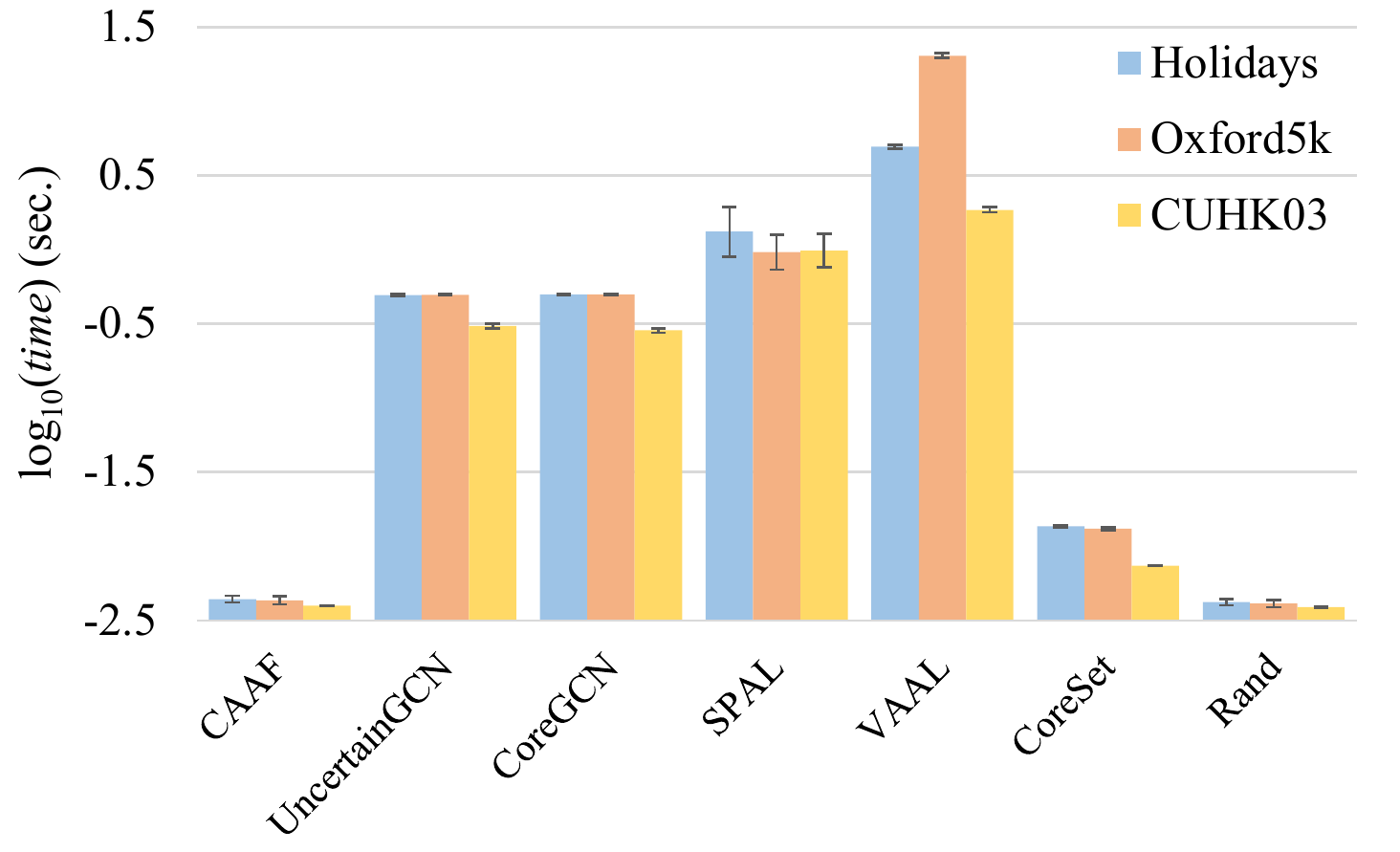}
  \caption{Execution time comparison of different AL methods on Holidays, Oxford5k and CUHK03.}
  \label{img-time}
\end{figure} 

\subsection{Comparison with Existing AL Methods}
\label{exp:comparison}

We compare CAAF with 6 AL methods on the three image-based datasets:
\begin{itemize}
  \item \textbf{Rand} is a commonly utilized baseline for AL approaches, where feedback samples are randomly chosen from the unlabeled set. 
  \item \textbf{CoreSet}~\cite{ICLR18-CoreSet} is a state-of-the-art geometric technique and it aims to choose a representative subset of the entire dataset. 
  \item \textbf{VAAL}~\cite{ICCV19-VAAL} aims to select the most representative samples by training a variational autoencoder and an adversarial network to discriminate between unlabeled data and labeled data. The probability associated with the discriminator's predictions serves as a score to select low-confidence samples for feedback.
  \item \textbf{SPAL}~\cite{AAAI19-SPAL} simultaneously considers the potential value and easiness of an instance by integrating a self-paced regularizer and a term that estimates the distribution difference between labeled data and unlabeled data into one objective function.
  \item \textbf{UncertainGCN}~\cite{CVPR21-SequentialGCN} constructs a sequential graph convolutional network (GCN) to distinguish unlabeled samples from labeled samples. The outputs of the GCN serve as confidence scores, and an uncertainty sampling approach is applied to select samples whose confidence scores are closest to a pre-defined small margin.
  \item \textbf{CoreGCN}~\cite{CVPR21-SequentialGCN} integrates CoreSet and UncertainGCN to select informative and representative feedback samples.
\end{itemize}

All baseline methods are implemented with their default settings in our experiment, except for VAAL, as it requires a relatively large number of labeled samples to train the VAE and adversarial network. However, it is impractical for our interactive retrieval settings, where the probe is the only labeled sample at the very beginning. Therefore, we randomly select $q$ samples in the first round of feedback, and VAAL is applied from the second round of feedback. For a fair comparison, all baseline methods, as with CAAF, are applied to the top-$K$ samples in the initial ranking list. In addition, these methods are only used to generate feedback suggestions for RF, and the feedback scores are diffused by the same ranking process described in Sec.~\ref{sec:ranking}. 

Figure~\ref{img-exp-main} shows the dynamic performance of all AL methods as the feedback round $T$ increases on three datasets. Compared with state-of-the-art AL methods, CAAF gains the highest improvement after 4 rounds of feedback, with the mAP improving more than 15\% on Holidays and Oxford5k and more than 45\% on CUHK03. Notably, different from the experimental results reported in other literature, some methods, \textit{e.g.}, CoreSet and VAAL, may be inferior to Rand, which can be attributed to the lack of labeled samples. These AL methods usually need a pre-defined labeled set that accounts for 5\%$\sim$10\% of the whole unlabeled set for a cold start, yet the query is the only labeled sample in the first round of feedback in interactive INS tasks. These results demonstrate that CAAF is more compatible with interactive INS tasks compared with state-of-the-art AL methods.

Figure~\ref{img-time} compares the mean and standard deviation of the execution time in each round of feedback on all three datasets. We observe that most baseline methods need more than 400 ms ($\approx \log_{10}(-0.39)$s) to generate one round of feedback suggestions, which cannot satisfy the optimal computer response time~\cite{Doherty}. However, with our approximate solution, CAAF is able to generate feedback suggestions in less than 5 ms, demonstrating its computational efficiency.

\subsection{Evaluation on NIST TRECVID}
In this section, we evaluate our method on the INS task of NIST TRECVID 2020 \& 2021 to demonstrate the capability of handling extremely large-scale datasets.

\subsubsection{Evaluation on NIST TRECVID 2021}
We report the official evaluation results~\cite{techreport} of the top-3 interactive runs and their corresponding automatic runs in 2021: 

\begin{itemize}
  \item \texttt{F\_M\_A\_B\_WHU\_NERCMS.21\_2} (Auto1) is the official first-ranked automatic result.
  \item \texttt{F\_M\_A\_B\_WHU\_NERCMS.21\_4} (Auto2) is the official second-ranked automatic result.
  \item \texttt{I\_M\_A\_B\_WHU\_NERCMS.21\_1} (Auto1+TopK) is the official first-ranked interactive result. This method takes Auto1 as the initial search result, asks the user to verify the initially top-ranked video segments, and removes the irrelevant segments from the search list.
  \item \texttt{I\_M\_A\_B\_WHU\_NERCMS.21\_5} (Auto2+TopK) is the official second-ranked interactive result. This method takes the same interaction strategy as Auto1+TopK, and the only difference is that it takes Auto2, instead of Auto1, as the initial search result.
  \item \texttt{I\_M\_A\_B\_WHU\_NERCMS.21\_3} (Auto2+CAAF) is the official third-ranked interactive result that refines Auto2 via the proposed CAAF. 
\end{itemize}

\begin{table}[t]
  \renewcommand{\arraystretch}{1.3}
  \caption{Comparison Between TopK and CAAF in NIST TRECVID 2021} 
  \label{tab:trecvid}
  \centering
  \begin{tabular}{|c||cc|cc||cc|}
    \hline
    \multirow{3}{*}{Topic} & \multicolumn{4}{c||}{Official} & \multicolumn{2}{c|}{Ours}\\
    \cline{2-7}
    & \multicolumn{2}{c|}{Auto2+TopK} & \multicolumn{2}{c||}{Auto2+CAAF} & \multicolumn{2}{c|}{Auto2+CAAF*} \\
    & \#FB & AP (\%) & \#FB & AP (\%) & \#FB & AP (\%) \\
    \hline \hline
    9319 &176  &59.8 &75  &60.0 &176 & 70.7 \\ 
    9320 &169  &79.0 &67  &77.2 &169 & 80.2 \\ 
    9321 &182  &65.3 &80  &64.1 &182 & 72.1 \\ 
    9322 & 97  &56.4 &60  &53.2 & 97 & 59.1 \\ 
    9323 & 90  &51.5 &66  &53.1 & 90 & 57.8 \\ 
    9324 & 91  &49.4 &61  &52.8 & 91 & 56.0 \\ 
    9325 &120  &71.9 &77  &69.7 &120 & 75.3 \\ 
    9326 &138  &74.5 &99  &73.1 &138 & 78.4 \\ 
    9327 &130  &74.0 &90  &74.0 &130 & 75.7 \\ 
    9328 & 98  &20.9 &86  &29.0 & 98 & 34.3 \\ 
    9329 &114  &21.5 &90  &21.6 &114 & 24.5 \\ 
    9330 & 97  &51.9 &73  &51.4 & 97 & 55.2 \\ 
    9331 &104  &39.8 &80  &39.9 &104 & 45.9 \\ 
    9332 &116  &51.6 &80  &52.4 &116 & 60.2 \\ 
    9333 & 77  &40.4 &82  &38.0 & 77 & 45.9 \\ 
    9334 & 79  &35.2 &83  &30.6 & 79 & 45.1 \\ 
    9335 & 60  & 6.5 &80  & 8.7 & 60 & 14.0 \\ 
    9336 & 67  &16.5 &96  &15.2 & 67 & 24.0 \\ 
    9337 & 50  &12.8 &72  &12.9 & 50 & 15.9 \\ 
    9338 & 53  &41.8 &70  &41.8 & 53 & 46.7 \\ 
    \hline \hline
    Mean &105.4&46.0 &78.4&45.9 &105.4& 51.9 \\
  \hline
  \end{tabular}
\end{table}

\begin{figure}[t]
  \centering
  \includegraphics[width = .9\linewidth]{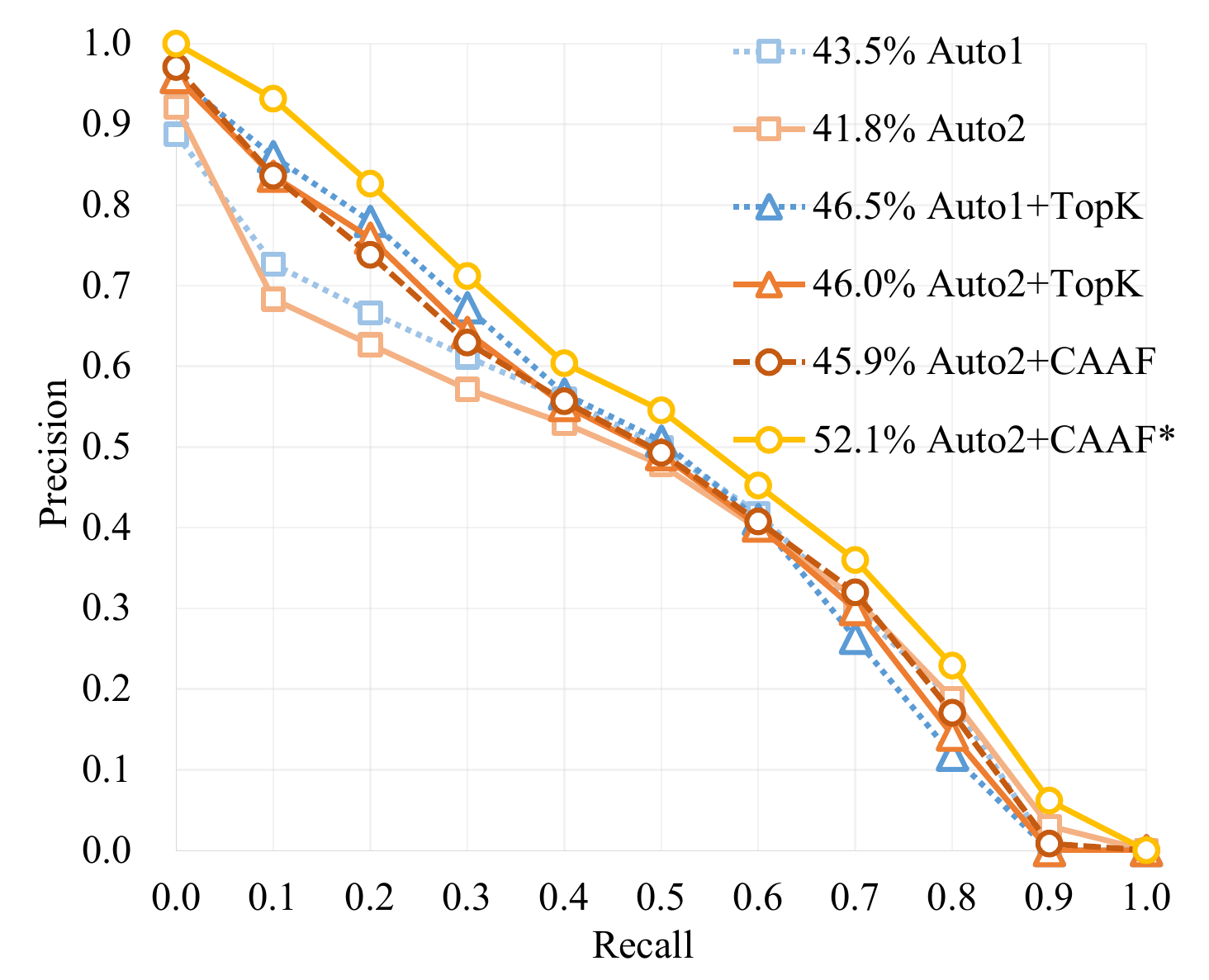}
  \caption{11-point interpolated precision-recall curve of Auto2+CAAF* and all baseline methods in NIST TRECVID 2021, where the values in the legend represent the mAP of each method.}\label{img-pr-curve}
\end{figure}

In addition to the official evaluation results, we supplement an interactive solution (Auto2+CAAF*) where the total feedback number of each topic is equivalent to that of Auto2+TopK, \textit{i.e.}, 176 samples are fed back to the user in Topic 9319, 169 samples are fed back to the user in Topic 9320, \textit{etc}.

As presented in Table~\ref{tab:trecvid}, when the total interaction time is constrained to 5 minutes, the user can check 105 video segments on average with Auto2+TopK, which is 28 more than that of Auto2+CAAF. The low-confidence samples selected by CAAF can sometimes be difficult for human users to discriminate. To alleviate the effects caused by noisy annotations, we ask the user to abstain from feedback if he or she is unsure about the label. Nevertheless, Auto2+TopK only performs 0.1\% better than Auto2+CAAF with the same initial search result. With both the same initial search result and total feedback number, Auto2+CAAF* outperforms Auto2+TopK by 5.9\%. Moreover, Figure~\ref{img-pr-curve} shows that Auto2+CAAF* even surpasses the first-ranked Auto1+TopK, although the performance of Auto2 itself is inferior to that of Auto1. \label{sec:trecvid}

All the above experimental results demonstrate that CAAF is capable of handling massive data and that CAAF is more efficient than directly checking the top-ranked samples in the initial ranking result.

\subsubsection{Evaluation on NIST TRECVID 2020}
We further implement Auto2+TopK and Auto2+CAAF on NIST TRECVID 2020. We fix the total feedback number of each topic to 80, as CAAF can check approximately 78 samples in 5 minutes according to the statistics in Table~\ref{tab:trecvid}. And we set $K=2000$, $q=16$ and $T=5$ for CAAF. The results are shown in Table~\ref{tab:trecvid20}. As a reference, we also report the best automatic and interactive results (denoted as ``AutoBest'' and ``AutoBest+InterBest'', respectively) on the official evaluation in 2020~\cite{PKU}.

We observe that although Auto2 is inferior to AutoBest, Auto2+TopK outperforms AutoBest+InterBest by 1.3\%, which means that TopK is quite a strong baseline. However, Auto2+CAAF outperforms Auto2+TopK by 0.6\%, which further demonstrates the effectiveness of CAAF.

\begin{table}[t]
  \renewcommand{\arraystretch}{1.3}
  \centering
  \caption{Evaluation on NIST TRECVID 2020}\label{tab:trecvid20}
  \begin{tabular}{|c|c|}
    \hline
    Method & mAP (\%)\\
    \hline
    AutoBest~\cite{PKU} & 25.2 \\
    AutoBest+InterBest~\cite{PKU} & 36.8 \\
    \hline
    Auto2 & 24.7 \\
    Auto2+TopK & 38.1 \\
    Auto2+CAAF & 38.7 \\
    \hline
  \end{tabular}
\end{table}

\subsection{Ablation Studies}

In this section, we examine the roles of two key modules, the \textit{ranking step} and \textit{suggestion step}, on the Oxford5k dataset by removing them from CAAF. When the \textit{ranking step} is removed, the ranking scores of the labeled samples are directly replaced by their feedback scores, while the ranking scores of the unlabeled samples retain their initial ranking scores. And when the \textit{suggestion step} is removed, the feedback suggestions are generated by random sampling. 

\begin{figure}[t]
  \centering
  \includegraphics[width = .85\linewidth]{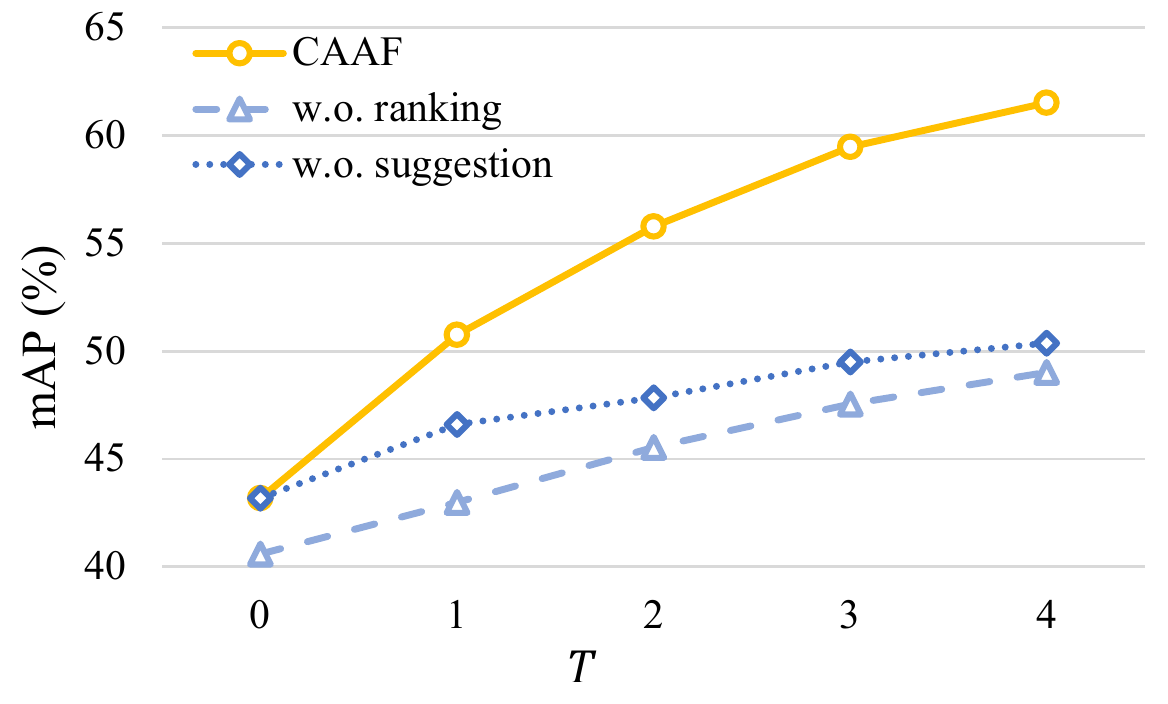}
  \caption{Ablation study on the Oxford5k dataset.}\label{img-ablation}
\end{figure}

We observe from Figure~\ref{img-ablation} that the performance of CAAF significantly drops once the \textit{ranking step} or \textit{suggestion step} is removed, which demonstrates that each module is essential to CAAF. Since the \textit{ranking step} can be regarded as modified MR, while the \textit{suggestion step} serves as an AL sampling strategy, it also implies that CAAF effectively takes advantage of both MR and AL.

\subsection{Analysis of Confidence Modeling}

\label{exp:confidence-modeling}

We claim in Section~\ref{sec:cwmr} that the confidence modeling scheme in CAAF can not only indicate valuable feedback samples for AL but also modulate the propagation weight in MR and improve the ranking accuracy. The comparative results in Section~\ref{exp:comparison} have already demonstrated the former part of the claim; this section is divided into two parts to further prove the claim. In the first part, we analyze the effect of weight modulation to demonstrate the latter part of the claim. In the second part, we further analyze the distribution of feedback samples and present an illustrative example of the feedback samples. The experiments are conducted on three image-based datasets with default settings.

\begin{table}[t]
  \renewcommand{\arraystretch}{1.3}
  \centering
  \caption{Comparison Between CAAF and MR}\label{tab:ablation}
  \begin{tabular}{|c|c|c|c|c|}
    \hline
    \multirow{2}{*}{$T$} & \multirow{2}{*}{Method} & Holidays & Oxford5k & CUHK03 \\
    & & mAP (\%) & mAP (\%) & mAP (\%) \\
    \hline
    \multirow{2}{*}{0}  & MR & \textbf{67.80} & \textbf{43.35} & 53.72 \\
                        & CAAF & 67.14 & 43.19 & \textbf{53.90} \\
    \hline
    \multirow{2}{*}{1}  & MR & 79.02 & 49.53 & 71.14 \\
                        & CAAF & \textbf{79.83} & \textbf{50.75} & \textbf{72.17}\\
    \hline
    \multirow{2}{*}{2}  & MR   & 82.30 & 53.51 & 83.87 \\
                        & CAAF & \textbf{84.02} & \textbf{55.78} & \textbf{85.55}\\
    \hline
    \multirow{2}{*}{3}  & MR   & 84.83 & 56.44 & 88.25 \\
                        & CAAF & \textbf{86.21} & \textbf{59.46} & \textbf{89.97}\\
    \hline
    \multirow{2}{*}{4}  & MR  & 86.29 & 58.39 & 90.60 \\
                        & CAAF & \textbf{87.46} & \textbf{61.51} & \textbf{91.33} \\
    \hline
  \end{tabular}
\end{table}

\subsubsection{Analysis of weight modulation}
To eliminate the effect of weight modulation in the ranking step, we set $\boldsymbol{v}=\boldsymbol{1}$ at the beginning of each round of feedback. In this case, the ranking step of CAAF degrades to the classical MR. We then compare CAAF and MR with the same suggestion step; the results are shown in Table~\ref{tab:ablation}.

As shown in Table~\ref{tab:ablation}, although CAAF performs slightly worse than MR when $T=0$, \textit{i.e.}, no sample is labeled except for the probe, it consistently outperforms MR on all three datasets when $T\geq 1$. The difference between CAAF and MR can reach 1.72\%, 3.12\%, and 1.72\% on Holidays, Oxford5k and CUHK03, respectively. These results demonstrate that $\boldsymbol{v}$ does help improve the ranking accuracy by modulating the affinity matrix in MR. 

\subsubsection{Analysis of feedback samples}
\begin{figure*}[!t]
  \centering
  \subfloat[Holidays]{\includegraphics[width = .3\linewidth]{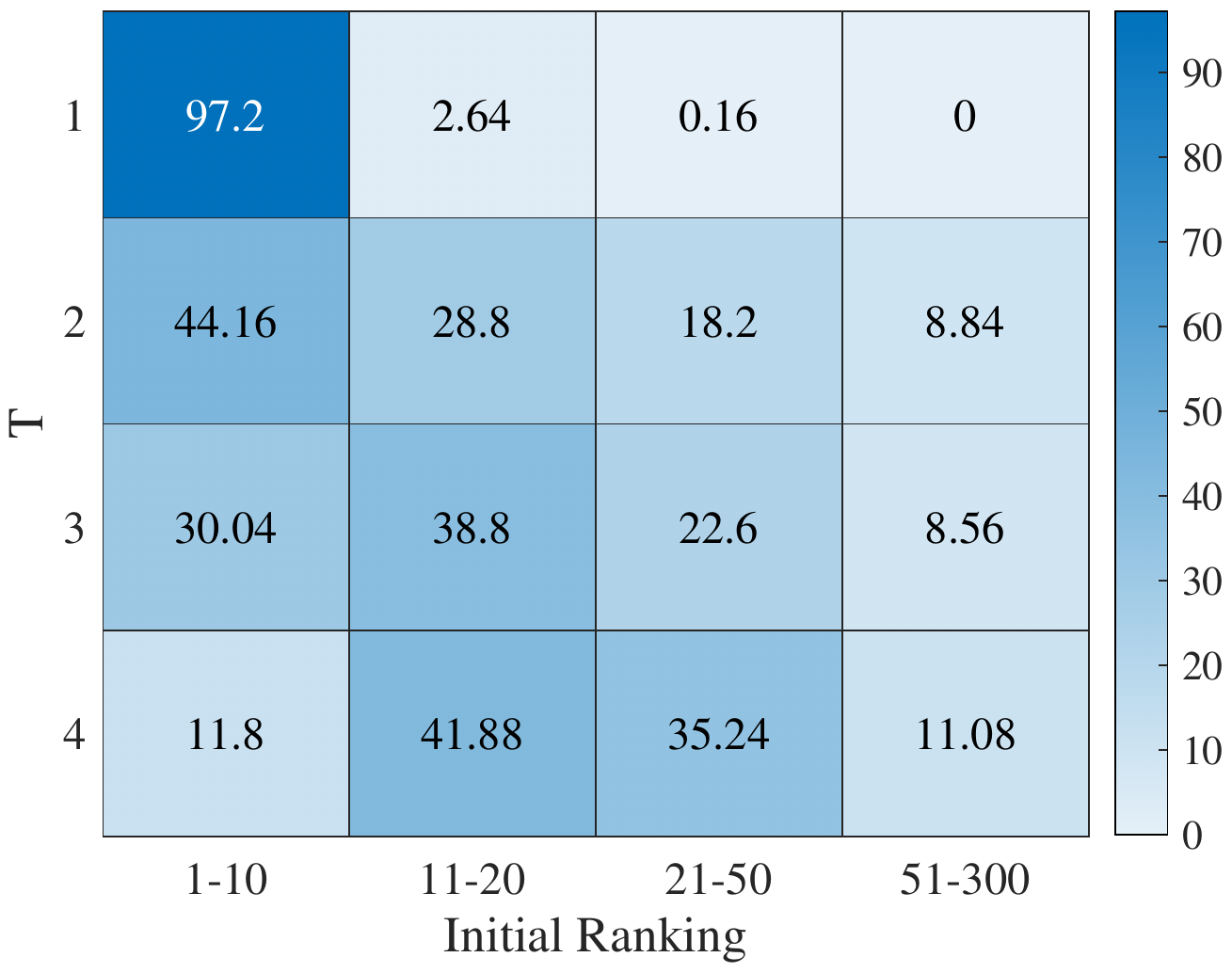}} 
  \hfil
  \subfloat[Oxford5k]{\includegraphics[width = .3\linewidth]{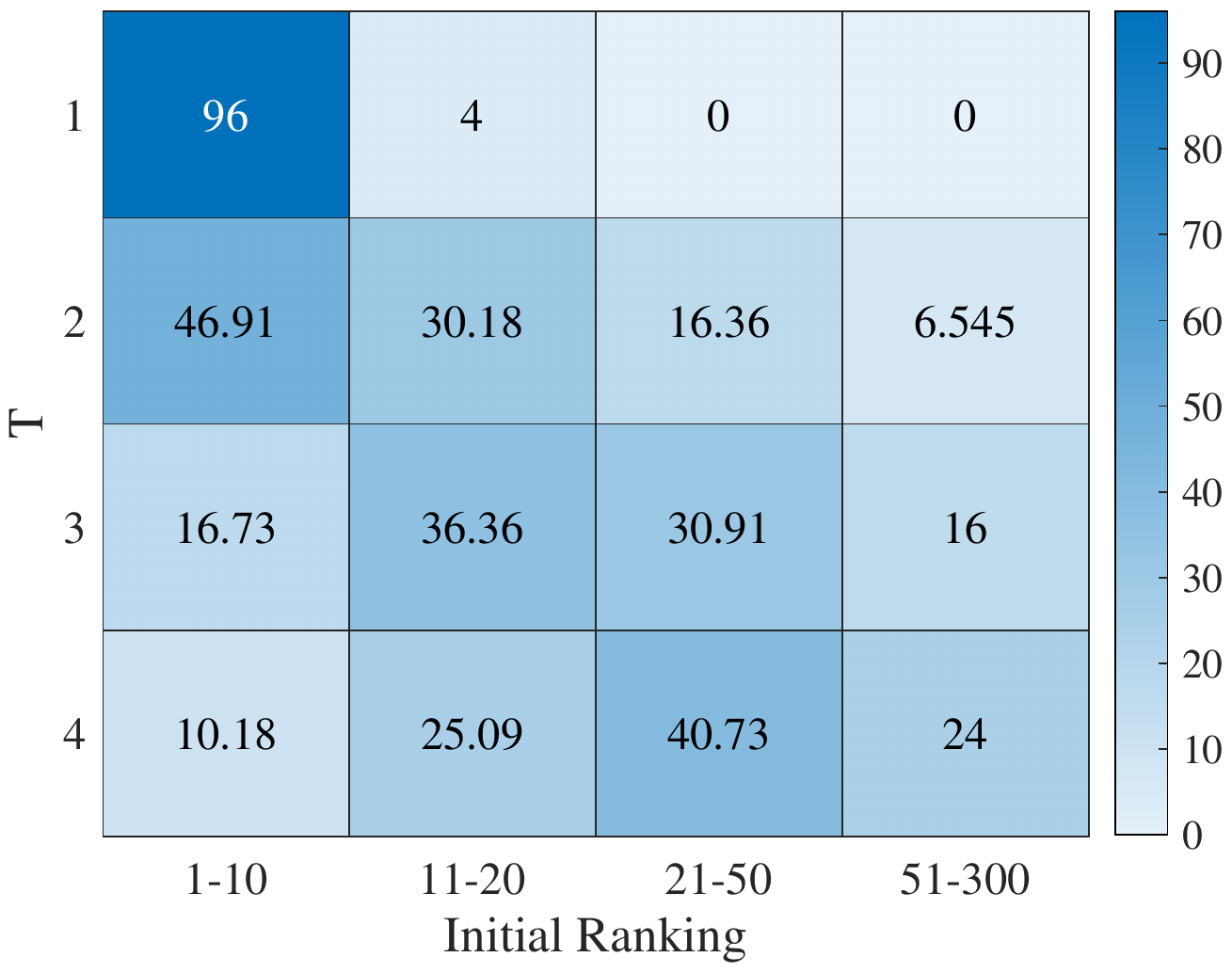}}
  \hfil
  \subfloat[CUHK03]{\includegraphics[width = .3\linewidth]{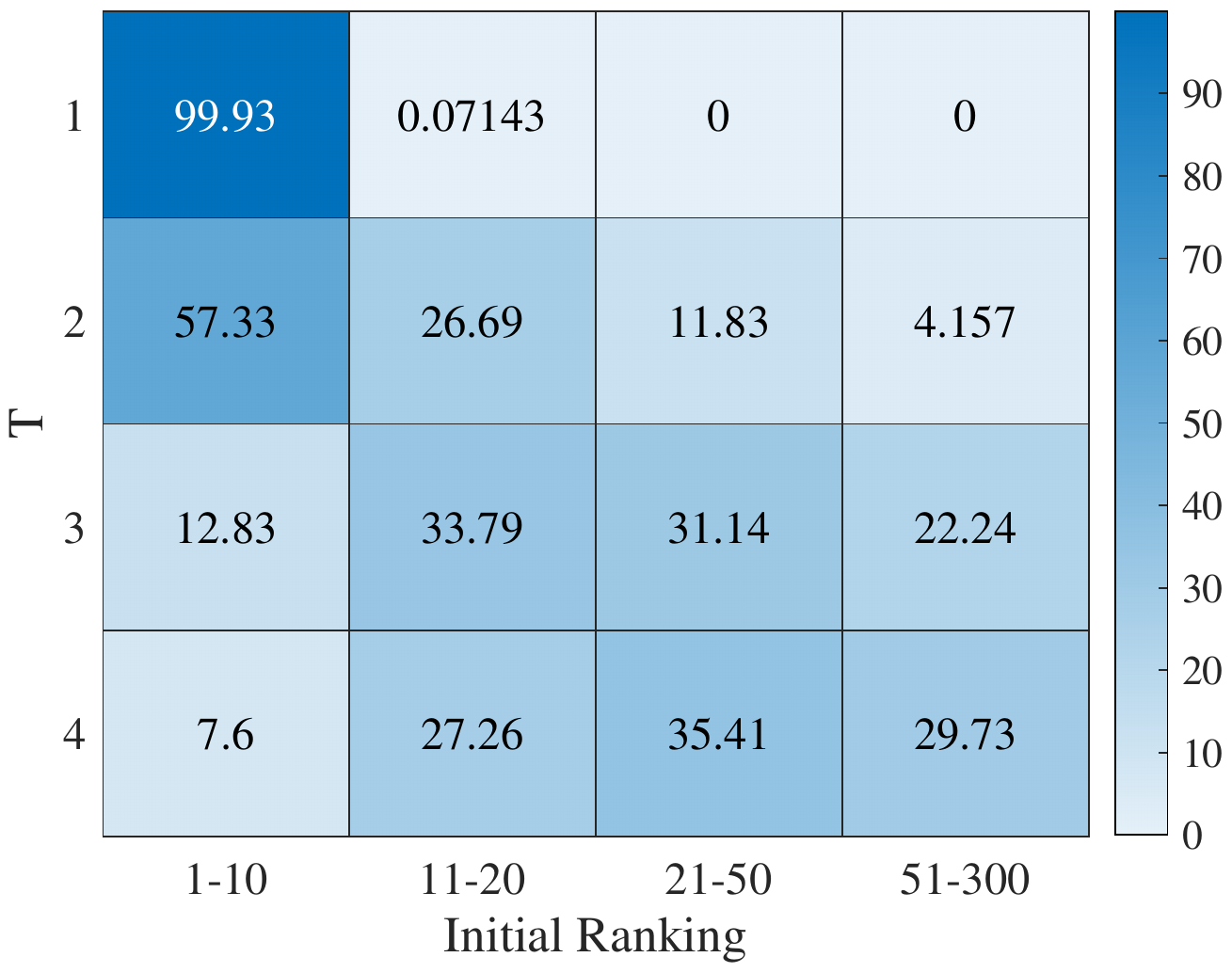}} 
  \hfil
  \caption{Heatmap of how feedback samples distribute in the initial ranking list, where the numerical values represent the proportion (\%) of feedback samples that are located at different initial ranking positions in each round of feedback, and the deeper color reflects the larger proportion.}\label{img-exp-heatmap}
\end{figure*}

We count the distribution of the feedback samples in the initial ranking list. The heatmap is illustrated in Figure~\ref{img-exp-heatmap}. We observe that in the first round of feedback, the feedback samples tend to be concentrated at the top of the initial ranking list. As $T$ increases, CAAF selects samples from the middle and back segments of the initial ranking list. This indicates that CAAF is able to select diverse feedback samples that are not so similar to the probe, which can be attributed to the manifold structure that can implicitly depict the data distribution. 

\begin{figure*}[!t]
  \centering
  \subfloat[Holidays]{\includegraphics[width = 0.3\linewidth]{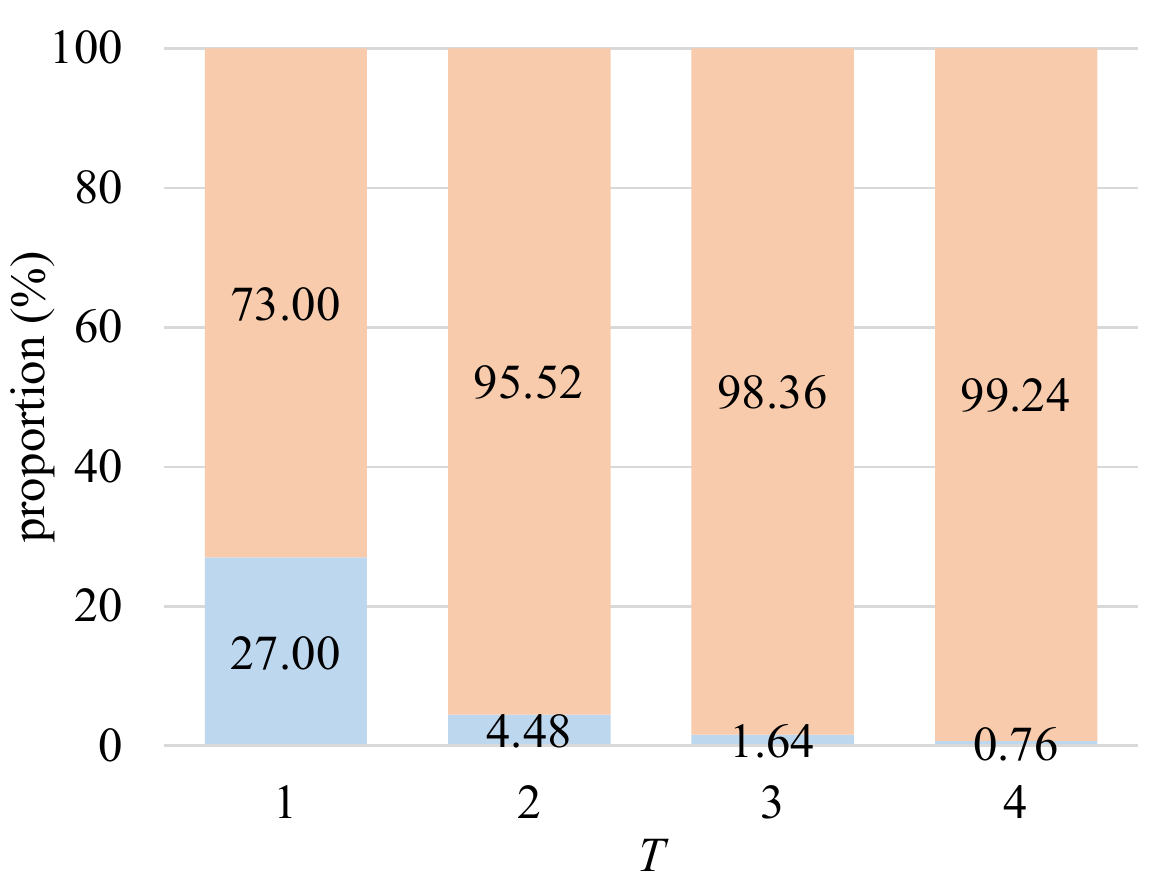} }
  \label{img-holiday-ratio}
  \hfil
  \subfloat[Oxford5k]{\includegraphics[width = 0.3\linewidth]{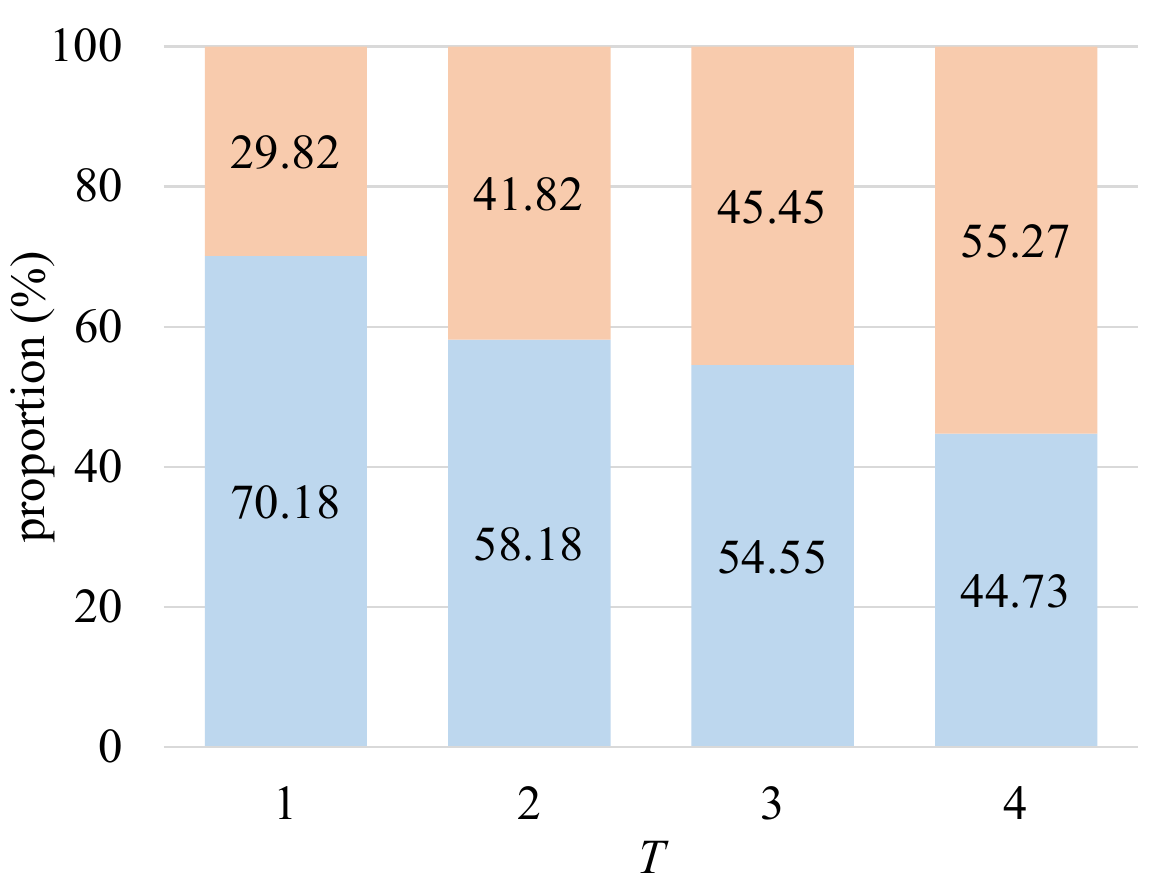} }
  \label{img-oxford-ratio}
  \hfil
  \subfloat[CUHK03]{\includegraphics[width = 0.3\linewidth]{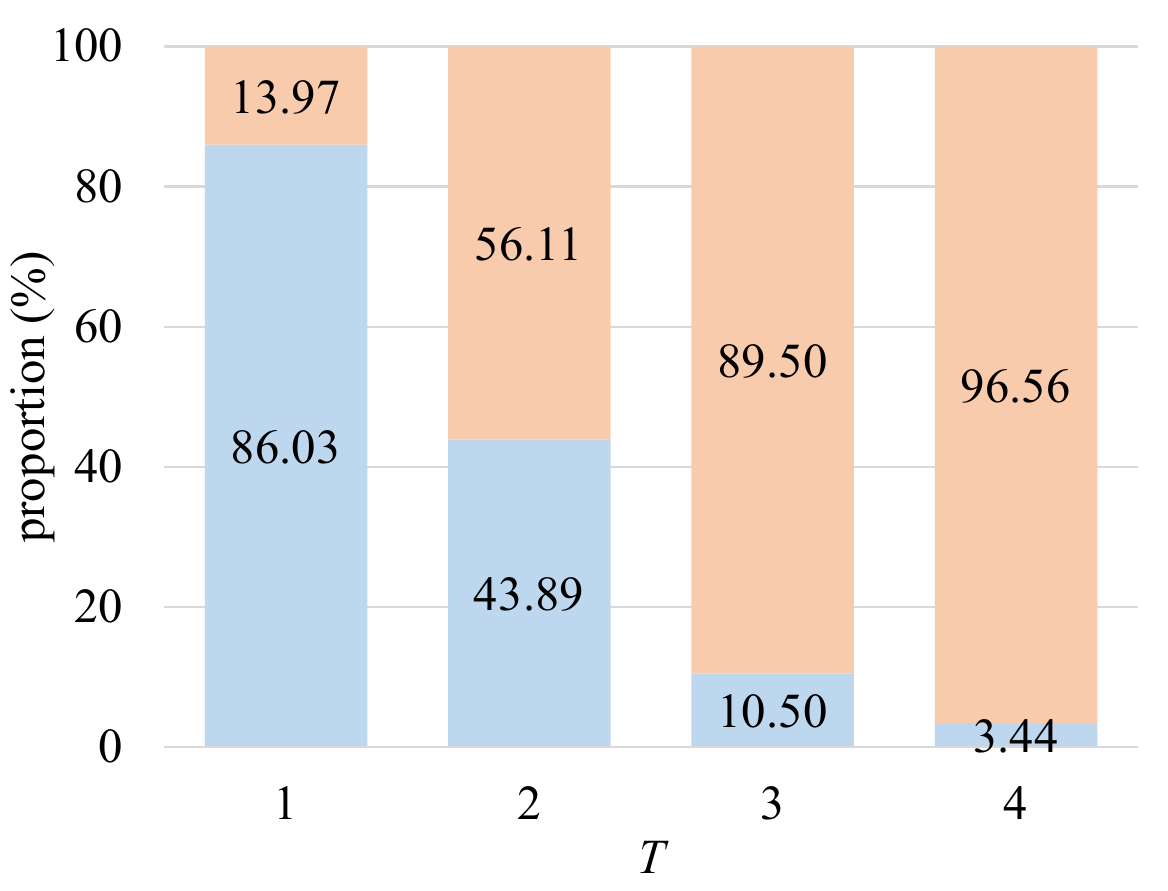} }
  \label{img-cuhk03-ratio}
  \hfil
  \subfloat{\includegraphics[width = 0.035\linewidth]{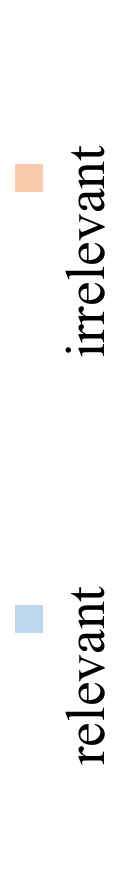}}
  \hfil
  \caption{Proportion of relevant and irrelevant feedback samples generated by CAAF, where the blue bars and orange bars represent the proportion (\%) of relevant samples and irrelevant samples, respectively, in each round of feedback.}\label{img-exp-ratio}
\end{figure*}

We further count the labels of the feedback samples. The proportion of relevant and irrelevant samples is illustrated in Figure~\ref{img-exp-ratio}. We observe that the proportion of relevant feedback samples gradually decreases as the round of feedback increases. Combined with Figure~\ref{img-exp-heatmap}, this finding indicates that relevant samples are more likely to appear at the top of the initial ranking list, demonstrating the rationality of selecting only the top-$K$ samples for interactive INS.

\begin{figure*}[!t]
  \centering
  \includegraphics[width = .85\linewidth]{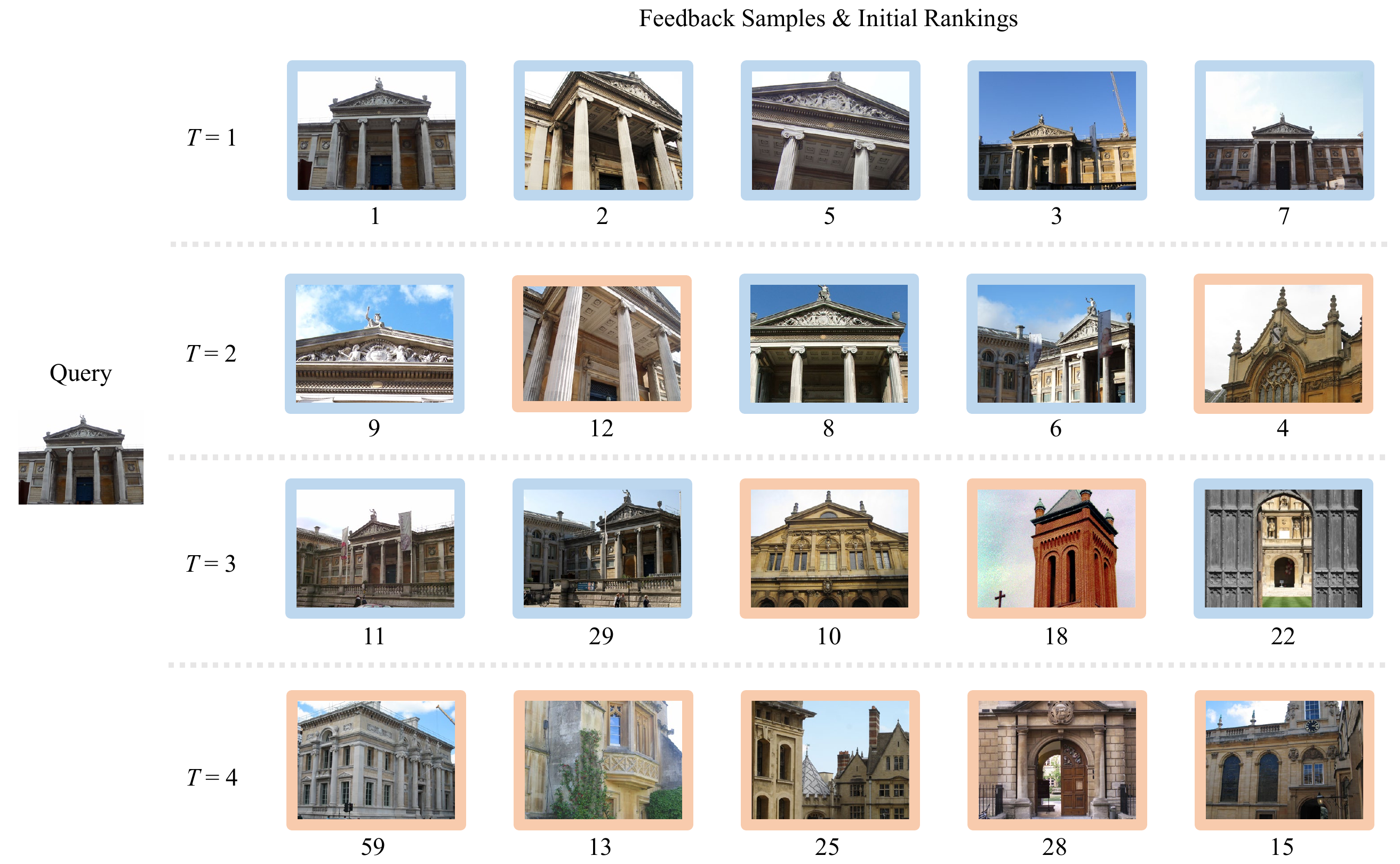}
  \caption{Visualized example of the feedback samples selected by CAAF on the Oxford5k dataset, where the first line of each row shows the feedback samples in the $T$-th round, and the second line shows their initial rankings. Images with blue bounding boxes and orange bounding boxes denote relevant samples and irrelevant samples, respectively.} \label{img-example}
\end{figure*}

A visualized example of the feedback samples selected by CAAF is shown in Figure~\ref{img-example}. As $T$ increases, the number of relevant samples gradually decreases, and so does the initial ranking of the selected samples.

\subsection{Analysis of Acceleration Strategies}\label{exp:rationality}
This section explores the rationality of the two acceleration strategies introduced in Section~\ref{sec:implementation} on the three image-based datasets with the default settings.

\begin{table}[!t]
  \arrayrulecolor{black}
  \renewcommand{\arraystretch}{1.3}
  \centering
  \caption{Comparison between QP Solution and Approximate Solution} \label{tab:why-approximate}
  \begin{tabular}{|c|c|cc|}
    \hline
    Dataset & Method & mAP (\%) & time (ms) \\ 
    \hline
    \multirow{2}{*}{Holidays} & QP & 87.60 & 50.8 $\pm$ 4.00\\ 
    & Appr. & 87.46 & {\color{white}0}4.3 $\pm$ 0.26 \\ 
    \hline
    \multirow{2}{*}{Oxford5k} & QP & 61.43 & 49.6 $\pm$ 0.10\\ 
    & Appr. & 61.51 & {\color{white}0}4.4 $\pm$ 0.22 \\ 
    \hline
    \multirow{2}{*}{CUHK03} & QP & 91.38 & 50.4 $\pm$ 2.10\\ 
    & Appr. & 91.33 & {\color{white}0}4.0 $\pm$ 0.02 \\ 
    \hline
  \end{tabular}
\end{table}

\subsubsection{Analysis of approximate solution}
We compare the search accuracy and execution time taken in each round of feedback of both the GPU-accelerated approximate solution (Appr.) and the original QP solution (QP); the results are shown in Table~\ref{tab:why-approximate}. We observe that Appr. achieves equivalent performance as QP; however, the former's execution time is only approximately one-tenth of the latter's. Therefore, it is rational to simplify the optimization process through the proposed approximation scheme.

\begin{figure}[!t]
    \centering
    \includegraphics[width = .9\linewidth]{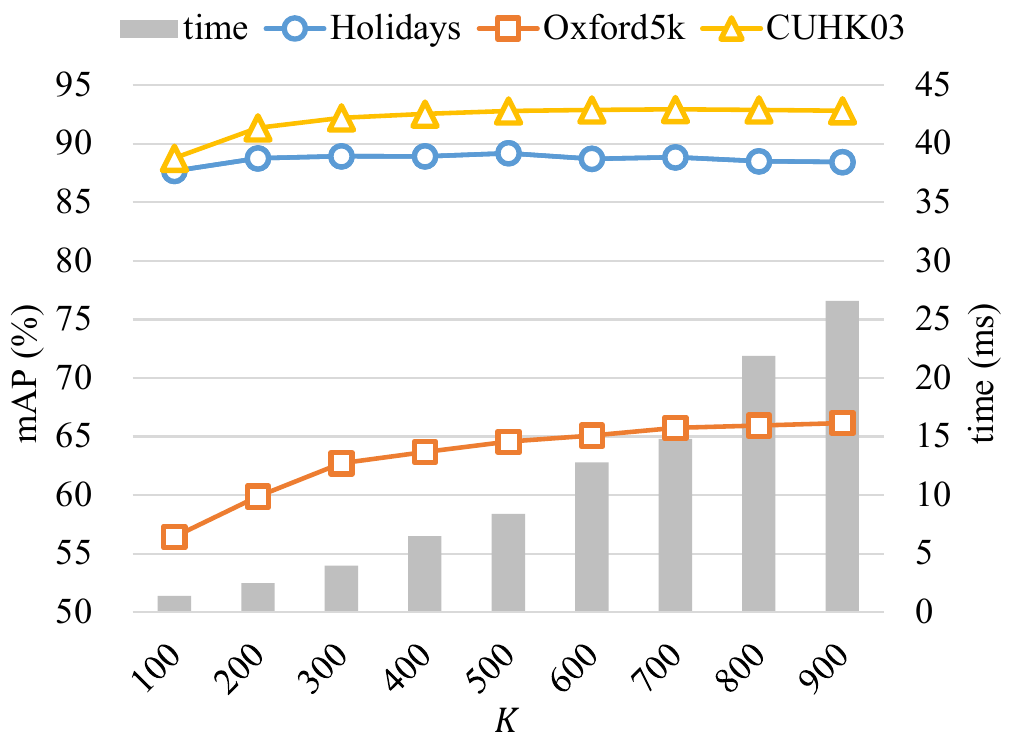}%
    \caption{Search accuracy and execution time with varying $K$s.}\label{img-k}
\end{figure}

\subsubsection{Analysis of top-$K$ search scheme}
The search accuracy and execution time with varying $K$s are illustrated in Figure~\ref{img-k}. We observe that the performance tends to be stable when $K\geq 300$. The execution time per round of query exponentially increases as $K$ increases. These results demonstrate that selecting only the top-$K$ galleries is a reasonable practice for balancing search accuracy and time cost, and we set $K=300$.

\begin{figure}[!t]
  \centering
  \includegraphics[width = .9\linewidth]{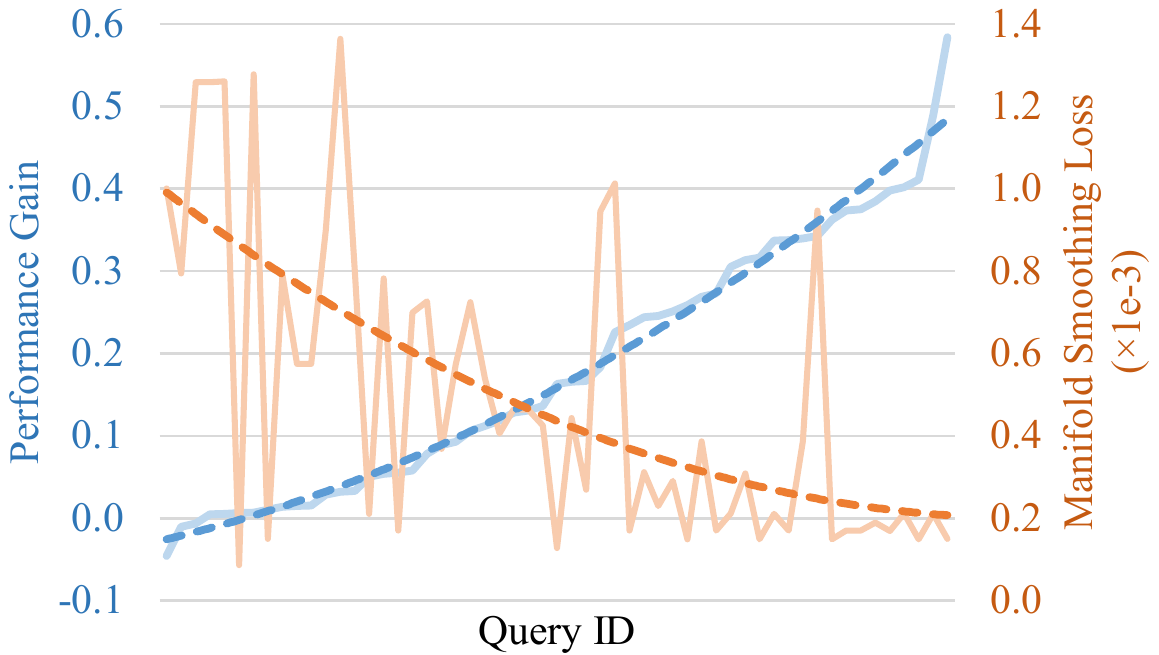}
  \caption{Performance gain/manifold smoothing loss \textit{v.s.} query ID on Oxford5k, where the blue line and orange line represent the performance gain and manifold smoothing loss, respectively, and the dashed lines are their second-order polynomial trend lines.}\label{img-limitations}
\end{figure}

\subsection{Limitations}
When analyzing the detailed performance gain of each query after 4 rounds of interaction with CAAF, we observe some cases where CAAF achieves minimal performance gain and even degradation. Since CAAF is built on the manifold ranking framework, the effectiveness of CAAF is limited when the data distribution does not conform to the manifold assumption. 

To validate this argument, we compute the manifold smoothing loss for each query with its ground truth label by $\frac{1}{m^2}\sum_{i=1}^{m}\sum_{j=1}^{m}a_{ij}(\hat{y}_i-\hat{y}_j)^2$, where $a_{ij}$ is the visual similarity between pairwise samples $x_i$ and $x_j$, and $\hat{y}_i=1$ if $x_i$ is the ground truth and $\hat{y}_i=0$, otherwise. Next, we rank each query's ID according to its performance gain in ascending order, and plot the corresponding manifold smoothing loss on Oxford5k. The results are shown in Figure~\ref{img-limitations}, where the blue line and orange line represent the performance gain and manifold smoothing loss, respectively, and the dashed lines denote their second-order polynomial trend lines. The figure shows that queries with higher loss values tend to obtain lower performance gain, which confirms the above argument.

\section{Conclusions}\label{sec:conclusions}
This paper investigates CAAF, a method that is specifically designed for interactive INS to improve the interaction efficiency by selecting the most valuable samples for RF. The core idea and main novelty lies in the explicit assessment of the ranking confidence, which improves not only the interaction efficiency by indicating valuable feedback candidates but also the retrieval performance by modulating the weights in the ranking loss. Furthermore, with an approximate solution and a top-$K$ search scheme, CAAF can be efficiently applied to interactive INS on large-scale datasets. Extensive experiments on both image INS tasks and video INS tasks demonstrate the effectiveness of our proposed method.

\section*{Acknowledgments}
Dr. Chao Liang would like to thank Prof. Philip S. Yu for his selfless help and valuable guidance about this work during Dr. Chao Liang's academic visit to UIC in 2018-2019. The numerical calculations in this paper have been done on the supercomputing system in the Supercomputing Center of Wuhan University. 

\bibliographystyle{IEEEtran}
\bibliography{manuscript}

\begin{thebibliography}{10}
\providecommand{\url}[1]{#1}
\csname url@samestyle\endcsname
\providecommand{\newblock}{\relax}
\providecommand{\bibinfo}[2]{#2}
\providecommand{\BIBentrySTDinterwordspacing}{\spaceskip=0pt\relax}
\providecommand{\BIBentryALTinterwordstretchfactor}{4}
\providecommand{\BIBentryALTinterwordspacing}{\spaceskip=\fontdimen2\font plus
\BIBentryALTinterwordstretchfactor\fontdimen3\font minus
  \fontdimen4\font\relax}
\providecommand{\BIBforeignlanguage}[2]{{%
\expandafter\ifx\csname l@#1\endcsname\relax
\typeout{** WARNING: IEEEtran.bst: No hyphenation pattern has been}%
\typeout{** loaded for the language `#1'. Using the pattern for}%
\typeout{** the default language instead.}%
\else
\language=\csname l@#1\endcsname
\fi
#2}}
\providecommand{\BIBdecl}{\relax}
\BIBdecl

\bibitem{TIP15-INS}
T.~Lin, M.~Yang, C.~Tsai, and Y.~F. Wang, ``Query-adaptive multiple instance
  learning for video instance retrieval,'' \emph{{IEEE} Trans. Image Process.},
  vol.~24, no.~4, pp. 1330--1340, 2015.

\bibitem{TMM16-INS}
J.~Meng, J.~Yuan, J.~Yang, G.~Wang, and Y.~Tan, ``Object instance search in
  videos via spatio-temporal trajectory discovery,'' \emph{{IEEE} Trans.
  Multim.}, vol.~18, no.~1, pp. 116--127, 2016.

\bibitem{TMM18-INS}
S.~D. Bhattacharjee, J.~Yuan, Y.~Huang, J.~Meng, and L.~Duan, ``Query adaptive
  multiview object instance search and localization using sketches,''
  \emph{{IEEE} Trans. Multim.}, vol.~20, no.~10, pp. 2761--2773, 2018.

\bibitem{TPAMI11-action}
J.~Yuan, Z.~Liu, and Y.~Wu, ``Discriminative video pattern search for efficient
  action detection,'' \emph{{IEEE} Trans. Pattern Anal. Mach. Intell.},
  vol.~33, no.~9, pp. 1728--1743, 2011.

\bibitem{TIP20-action}
Y.~Liu, Z.~Lu, J.~Li, T.~Yang, and C.~Yao, ``Deep image-to-video adaptation and
  fusion networks for action recognition,'' \emph{{IEEE} Trans. Image
  Process.}, vol.~29, pp. 3168--3182, 2020.

\bibitem{TIP21-action}
Y.~Liu, K.~Wang, G.~Li, and L.~Lin, ``Semantics-aware adaptive knowledge
  distillation for sensor-to-vision action recognition,'' \emph{{IEEE} Trans.
  Image Process.}, vol.~30, pp. 5573--5588, 2021.

\bibitem{TIP21-hashing}
W.~Zhao, Z.~Guan, H.~Luo, J.~Peng, and J.~Fan, ``Deep multiple instance hashing
  for fast multi-object image search,'' \emph{{IEEE} Trans. Image Process.},
  vol.~30, pp. 7995--8007, 2021.

\bibitem{TMM21-metric}
X.~Yao, D.~She, H.~Zhang, J.~Yang, M.~Cheng, and L.~Wang, ``Adaptive deep
  metric learning for affective image retrieval and classification,''
  \emph{{IEEE} Trans. Multim.}, vol.~23, pp. 1640--1653, 2021.

\bibitem{TMM18-RF}
Z.~Chen, Z.~Xu, Y.~Zhang, and X.~Gu, ``Query-free clothing retrieval via
  implicit relevance feedback,'' \emph{{IEEE} Trans. Multim.}, vol.~20, no.~8,
  pp. 2126--2137, 2018.

\bibitem{KBS21-RF}
M.~Zhao, J.~Liu, Z.~Zhang, and J.~Fan, ``A scalable sub-graph regularization
  for efficient content based image retrieval with long-term relevance feedback
  enhancement,'' \emph{Knowl. Based Syst.}, vol. 212, p. 106505, 2021.

\bibitem{TR09-AL}
B.~Settles, ``{Active learning literature survey},'' University of
  Wisconsin-Madison Department of Computer Sciences, Tech. Rep., 2009.

\bibitem{ICLR18-CoreSet}
O.~Sener and S.~Savarese, ``Active learning for convolutional neural networks:
  {A} core-set approach,'' in \emph{6th International Conference on Learning
  Representations}, 2018.

\bibitem{ICCV19-VAAL}
S.~Sinha, S.~Ebrahimi, and T.~Darrell, ``Variational adversarial active
  learning,'' in \emph{IEEE International Conference on Computer Vision}, 2019,
  pp. 5971--5980.

\bibitem{AAAI19-SPAL}
Y.~Tang and S.~Huang, ``Self-paced active learning: Query the right thing at
  the right time,'' in \emph{The Thirty-Third AAAI Conference on Artificial
  Intelligence}, 2019, pp. 5117--5124.

\bibitem{CVPR21-SequentialGCN}
R.~Caramalau, B.~Bhattarai, and T.~Kim, ``Sequential graph convolutional
  network for active learning,'' in \emph{{IEEE} Conference on Computer Vision
  and Pattern Recognition}, 2021, pp. 9583--9592.

\bibitem{NIPS10-SPL}
M.~P. Kumar, B.~Packer, and D.~Koller, ``Self-paced learning for latent
  variable models,'' in \emph{Advances in Neural Information Processing
  Systems}, 2010, pp. 1189--1197.

\bibitem{NIPS14-SPLD}
L.~Jiang, D.~Meng, S.~Yu, Z.~Lan, S.~Shan, and A.~G. Hauptmann, ``Self-paced
  learning with diversity,'' in \emph{Advances in Neural Information Processing
  Systems}, 2014, pp. 2078--2086.

\bibitem{NIPS03-MR}
D.~Zhou, J.~Weston, A.~Gretton, O.~Bousquet, and B.~Sch{\"{o}}lkopf, ``Ranking
  on data manifolds,'' in \emph{Advances in Neural Information Processing
  Systems}, 2003, pp. 169--176.

\bibitem{TPAMI18-ASPL}
L.~Lin, K.~Wang, D.~Meng, W.~Zuo, and L.~Zhang, ``Active self-paced learning
  for cost-effective and progressive face identification,'' \emph{{IEEE} Trans.
  Pattern Anal. Mach. Intell.}, vol.~40, no.~1, pp. 7--19, 2018.

\bibitem{alt-opt}
J.~C. Bezdek and R.~J. Hathaway, ``Convergence of alternating optimization,''
  \emph{Neural, Parallel \& Scientific Computations}, vol.~11, no.~4, pp.
  351--368, 2003.

\bibitem{TPAMI18-INS-Survey}
L.~Zheng, Y.~Yang, and Q.~Tian, ``{SIFT} meets {CNN:} {A} decade survey of
  instance retrieval,'' \emph{{IEEE} Trans. Pattern Anal. Mach. Intell.},
  vol.~40, no.~5, pp. 1224--1244, 2018.

\bibitem{TMM21-videoretrieval}
L.~Rossetto, R.~Gasser, J.~Lokoc, W.~Bailer, K.~Schoeffmann, B.~M{\"{u}}nzer,
  T.~Soucek, P.~A. Nguyen, P.~Bolettieri, A.~Leibetseder, and S.~Vrochidis,
  ``Interactive video retrieval in the age of deep learning - detailed
  evaluation of {VBS} 2019,'' \emph{{IEEE} Trans. Multim.}, vol.~23, pp.
  243--256, 2021.

\bibitem{TMM22-reid}
Y.~Li, H.~Yao, and C.~Xu, ``Intra-domain consistency enhancement for
  unsupervised person re-identification,'' \emph{{IEEE} Trans. Multim.},
  vol.~24, pp. 415--425, 2022.

\bibitem{TMM19-QE}
S.~Pang, J.~Ma, J.~Zhu, J.~Xue, and Q.~Tian, ``Improving object retrieval
  quality by integration of similarity propagation and query expansion,''
  \emph{{IEEE} Trans. Multim.}, vol.~21, no.~3, pp. 760--770, 2019.

\bibitem{Access21-RF}
H.~Ma, J.~Hou, C.~Zhu, W.~Zhang, R.~Tang, J.~Lai, J.~Zhu, X.~He, and Y.~Yu,
  ``{QA4PRF:} {A} question answering based framework for pseudo relevance
  feedback,'' \emph{{IEEE} Access}, vol.~9, pp. 139\,303--139\,314, 2021.

\bibitem{CVPR07-Oxford5k}
J.~Philbin, O.~Chum, M.~Isard, J.~Sivic, and A.~Zisserman, ``Object retrieval
  with large vocabularies and fast spatial matching,'' in \emph{{IEEE}
  Conference on Computer Vision and Pattern Recognition}, 2007.

\bibitem{ECCV08-Holidays}
H.~J{\'{e}}gou, M.~Douze, and C.~Schmid, ``Hamming embedding and weak geometric
  consistency for large scale image search,'' in \emph{European Conference on
  Computer Vision}, vol. 5302, 2008, pp. 304--317.

\bibitem{CVPR14-CUHK03}
W.~Li, R.~Zhao, T.~Xiao, and X.~Wang, ``Deepreid: Deep filter pairing neural
  network for person re-identification,'' in \emph{{IEEE} Conference on
  Computer Vision and Pattern Recognition}, 2014, pp. 152--159.

\bibitem{techreport}
Y.~Niu, J.~Yang, A.~Lu, B.~Huang, Y.~Zhang, J.~Huang, S.~Wen, D.~Xu, C.~Liang,
  Z.~Wang, and J.~Chen, ``{WHU-NERCMS} {AT} {TRECVID2021:} {INSTANCE} {SEARCH}
  {TASK},'' in \emph{2021 {TREC} Video Retrieval Evaluation}, 2021.

\bibitem{MTA17-AL4Retrieval}
X.~Zhao and G.~Ding, ``Query expansion for object retrieval with active
  learning using bow and {CNN} feature,'' \emph{Multim. Tools Appl.}, vol.~76,
  no.~9, pp. 12\,133--12\,147, 2017.

\bibitem{ICMLA18-DeepBASS}
M.~Rottmann, K.~Kahl, and H.~Gottschalk, ``Deep bayesian active semi-supervised
  learning,'' in \emph{Proceedings of the 17th International Conference on
  Machine Learning and Applications}, 2018, pp. 158--164.

\bibitem{ICCV19-DRAL}
Z.~Liu, J.~Wang, S.~Gong, D.~Tao, and H.~Lu, ``Deep reinforcement active
  learning for human-in-the-loop person re-identification,'' in
  \emph{{IEEE/CVF} International Conference on Computer Vision}, 2019, pp.
  6121--6130.

\bibitem{CVPR12-RALF}
S.~Ebert, M.~Fritz, and B.~Schiele, ``{RALF:} {A} reinforced active learning
  formulation for object class recognition,'' in \emph{{IEEE} Conference on
  Computer Vision and Pattern Recognition}, 2012, pp. 3626--3633.

\bibitem{TIP21-SPL-arcface}
X.~Wu, J.~Chang, Y.~Lai, J.~Yang, and Q.~Tian, ``Bispl: Bidirectional
  self-paced learning for recognition from web data,'' \emph{{IEEE} Trans.
  Image Process.}, vol.~30, pp. 6512--6527, 2021.

\bibitem{ACMMM21-SPL-triplet}
J.~Wei, X.~Xu, Z.~Wang, and G.~Wang, ``Meta self-paced learning for cross-modal
  matching,'' in \emph{Proceedings of the 29th ACM International Conference on
  Multimedia}.\hskip 1em plus 0.5em minus 0.4em\relax {ACM}, 2021, pp.
  3835--3843.

\bibitem{IJCV19-SPL}
D.~Zhang, J.~Han, L.~Zhao, and D.~Meng, ``Leveraging prior-knowledge for weakly
  supervised object detection under a collaborative self-paced curriculum
  learning framework,'' \emph{Int. J. Comput. Vis.}, vol. 127, no.~4, pp.
  363--380, 2019.

\bibitem{TPAMI20-SPL}
D.~Zhang, J.~Han, L.~Yang, and D.~Xu, ``{SPFTN:} {A} joint learning framework
  for localizing and segmenting objects in weakly labeled videos,''
  \emph{{IEEE} Trans. Pattern Anal. Mach. Intell.}, vol.~42, no.~2, pp.
  475--489, 2020.

\bibitem{NIPS20-SPL}
D.~Zhang, H.~Tian, and J.~Han, ``Few-cost salient object detection with
  adversarial-paced learning,'' in \emph{Advances in Neural Information
  Processing Systems}, 2020.

\bibitem{convex-opt}
S.~P. Boyd, L.~Vandenberghe, S.~P. Boyd, and L.~Vandenberghe, \emph{{Convex
  optimization}}.\hskip 1em plus 0.5em minus 0.4em\relax Cambridge university
  press, 2004.

\bibitem{SIGIR98-why-top-k}
J.~Zobel, ``How reliable are the results of large-scale information retrieval
  experiments?'' in \emph{Proceedings of the 21th Annual International ACM
  SIGIR Conference on Research and Development in Information Retrieval}, 1998,
  pp. 307--314.

\bibitem{SIGIR07-LOD}
X.~He, W.~Min, D.~Cai, and K.~Zhou, ``Laplacian optimal design for image
  retrieval,'' in \emph{Proceedings of the 30th Annual International ACM SIGIR
  Conference on Research and Development in Information Retrieval}, 2007, pp.
  119--126.

\bibitem{PRCV18-reid}
C.~Han, K.~Chen, J.~Wang, C.~Gao, and N.~Sang, ``Re-ranking person
  re-identification with adaptive hard sample mining,'' in \emph{Chinese
  Conference on Pattern Recognition and Computer Vision}, vol. 11256, 2018, pp.
  3--14.

\bibitem{CVPR17-k-reciprocal}
Z.~Zhong, L.~Zheng, D.~Cao, and S.~Li, ``Re-ranking person re-identification
  with k-reciprocal encoding,'' in \emph{{IEEE} Conference on Computer Vision
  and Pattern Recognition}, 2017, pp. 3652--3661.

\bibitem{ICLR15-VGG}
K.~Simonyan and A.~Zisserman, ``Very deep convolutional networks for
  large-scale image recognition,'' in \emph{3rd International Conference on
  Learning Representations}, 2015.

\bibitem{CVPR09-ImageNet}
J.~Deng, W.~Dong, R.~Socher, L.~Li, K.~Li, and F.~Li, ``Imagenet: {A}
  large-scale hierarchical image database,'' in \emph{{IEEE} Conference on
  Computer Vision and Pattern Recognition}, 2009, pp. 248--255.

\bibitem{ECCV18-PCB}
Y.~Sun, L.~Zheng, Y.~Yang, Q.~Tian, and S.~Wang, ``Beyond part models: Person
  retrieval with refined part pooling (and {A} strong convolutional
  baseline),'' in \emph{European Conference on Computer Vision}, vol. 11208,
  2018, pp. 501--518.

\bibitem{auto}
J.~Yang, C.~Liang, Y.~Niu, B.~Huang, and Z.~Wang, ``A spatio-temporal identity
  verification method for person-action instance search in movies,''
  \emph{CoRR}, vol. abs/2111.00228, 2021.

\bibitem{Doherty}
W.~J. Doherty and A.~J. Thadhani, ``The economic value of rapid response
  time,'' \emph{IBM Report}, 1982.

\bibitem{PKU}
Y.~Peng, Z.~Ye, J.~Zhang, and H.~Sun, ``Pku{\_}wict at {TRECVID} 2020: Instance
  search task,'' in \emph{2020 {TREC} Video Retrieval Evaluation}, 2020.

\end{thebibliography}

\vfill

\end{document}